\documentclass{IEEEtran}
\usepackage{amssymb,graphicx,color,booktabs,cite,multirow,multicol,amsfonts,textcomp,changepage,epstopdf,psfrag,algorithm,algorithmic,diagbox,mdwmath,bm,enumerate,colortbl,ragged2e,xcolor,amsmath,times,tikz,mathptmx}

\usepackage[hidelinks]{hyperref}
\usepackage{mathrsfs}
\usepackage[tight,footnotesize,center]{subfigure}
\usepackage{arydshln}
\definecolor{lime}{HTML}{A6CE39}
\DeclareRobustCommand{\orcidicon}{%
	\begin{tikzpicture}
	\draw[lime, fill=lime] (0,0) 
	circle [radius=0.16] 
	node[white] {{\fontfamily{qag}\selectfont \tiny ID}};
	\draw[white, fill=white] (-0.0625,0.095) 
	circle [radius=0.007];
	\end{tikzpicture}
	\hspace{-2mm}
}

\foreach \x in {A, ..., Z}{
	\expandafter\xdef\csname orcid\x\endcsname{\noexpand\href{https://orcid.org/\csname orcidauthor\x\endcsname}{\noexpand\orcidicon}}
}

\begin{document}
\title{\textsc{FusDreamer}: Label-efficient Remote Sensing World Model for Multimodal Data Classification}
\author{Jinping Wang\orcidA{},~\IEEEmembership{Member,~IEEE,} Weiwei Song\orcidD{},~Hao Chen\orcidE{},\\~Jinchang Ren\orcidG{},~\IEEEmembership{Senior Member,~IEEE,}~and~Huimin Zhao\orcidF{},~\IEEEmembership{Member,~IEEE}
\thanks{This work was supported by the Key Discipline Research Capacity Improvement Project of Guangdong Province under Grand 2024ZDJS022. (Corresponding author: Huimin Zhao.)}

\thanks{Jinping Wang and Huimin Zhao are with the School of Computer Sciences, Guangdong Polytechnic Normal University, Guangzhou, 510665, China, and also with the Guangdong Provincial Key Laboratory of Intellectual Property and Big Data, Guangdong Polytechnic Normal University, Guangzhou, 510665, China (e-mail: wangjp@gpnu.edu.cn; zhaohuimin@gpnu.edu.cn).

Weiwei Song is with the Peng Cheng Laboratory, Shenzhen, 518000, China. (e-mail: weiweisong415@gmail.com).

Hao Chen is with the Department of Applied Mathematics and Theoretical Physics, University of Cambridge, Cambridge, CB3 0WA, U.K. (e-mail: hc666@cam.ac.uk).

Jinchang Ren is with the School of Computer Sciences, Guangdong Polytechnic Normal University, Guangzhou 510640, China, and also with the National Subsea Centre, School of Computing, Engineering and Technology, Robert Gordon University, AB10 7AQ Aberdeen, U.K. (e-mail: jinchang.ren@ieee.org).
}}

\markboth{IEEE Transactions on Geoscience and Remote Sensing, ~Vol.~X, No.~X, 2024}
{Shell \MakeLowercase{\textit{et al.}}:}

\maketitle
\begin{abstract}
World models significantly enhance hierarchical understanding, improving data integration and learning efficiency. To explore the potential of the world model in the remote sensing (RS) field, this paper proposes a label-efficient remote sensing world model for multimodal data fusion (FusDreamer).
The FusDreamer uses the world model as a unified representation container to abstract common and high-level knowledge, promoting interactions across different types of data, \emph{i.e.}, hyperspectral (HSI), light detection and ranging (LiDAR), and text data. Initially, a new latent diffusion fusion and multimodal generation paradigm (LaMG) is utilized for its exceptional information integration and detail retention capabilities. Subsequently, an open-world knowledge-guided consistency projection (OK-CP) module incorporates prompt representations for visually described objects and aligns language-visual features through contrastive learning. {In this way, the domain gap can be bridged by fine-tuning the pre-trained world models with limited samples.} Finally, an end-to-end multitask combinatorial optimization (MuCO) strategy can capture slight feature bias and constrain the diffusion process in a collaboratively learnable direction. Experiments conducted on four typical datasets indicate the effectiveness and advantages of the proposed FusDreamer. The corresponding code will be released at \url{https://github.com/Cimy-wang/FusDreamer}.
\end{abstract}


\section{Introduction}
\IEEEPARstart{M}{ultimodal} data fusion, especially the hyperspectral (HSI), and light detection and ranging (LiDAR) data fusion, has attracted extensive attention in aboveground biomass estimation, city classification, and environmental monitoring \cite{10417796,yao2023laplacian}. In the remote sensing (RS) field, HSI and LiDAR provide distinct yet complementary visual spectrum, enhancing sensing capabilities and compensating for each other's limitations.

Recently, deep learning (DL) technology has significantly impacted the remote sensing field, using multiple convolutions, pooling, and fully connected layers to represent the relationships within the data \cite{li2019deep,zhang2024bipartite,Zhao_2023_CVPR}. For instance, Ge \emph{et al.} propose a deep residual fusion network, using the concated deep features from individual sources for HSI and LiDAR data classification \cite{ge2021deep}. Then, various improved convolution techniques, such as the dilated convolution \cite{pan2020dssnet}, deformable convolution \cite{WANG202247}, and orthogonal convolution \cite{huang2021graph}, have been developed to enhance feature representation. These techniques are generally better adapted to diverse data characteristics, thereby improving the efficiency and accuracy of channel-wise and spatial feature representation. Later, a vision transformer (ViT) is introduced as a groundbreaking architecture for image processing \cite{dosovitskiy2020image}. Compared with the traditional convolution neural networks (CNN), ViTs leverage self-attention mechanisms to effectively capture global dependencies within visual data, therefore possessing stronger feature capture and generalization abilities than CNNs. Various ViT-variants have catalyzed further advancements in the field of image recognition and classification, such as the Swin transformer \cite{liu2021swin} and ScalableViT \cite{yang2022scalablevit}. However, both CNNs and ViTs-based structures require a large amount of well-labeled training data to be fully trained, and it is hard for them to achieve optimal performance under a few labeled training samples \cite{10517881}. 

Researchers have demonstrated that models pre-trained on large datasets like ImageNet can be fine-tuned for specific remote sensing tasks, offering a potential solution to label-limited problems \cite{lee2022exploring,wang2023dcn}. For instance, Lee \emph{et al.} develop a cross-domain pre-trained model on the ImageNet dataset, extensively using the cross-domain approach with ones trained from scratch \cite{lee2022exploring}. However, directly transferring ImageNet pre-trained models to RS classification encounters a substantial domain gap due to significant differences between natural and RS images. Furthermore, both vision-centric foundation models only focus on specific vision pattern features, neglecting the semantic understanding of the objects and their relationships. For example, when conducting land cover classification, a vision-centric model might mistakenly classify a building rooftop pixel as a highway road if the pixel visually resembles a highway road \cite{li2024vision}.
Nowadays, the success of large language models has inspired extensive research into vision-language models (VLMs) \cite{zhang2024vision,Zhao_2024_ICML}. {To better utilize VLMs for RS data analysis, an important step is to integrate RS expert knowledge into VLMs properly, \emph{i.e.}, empowering large language models with domain-specific knowledge, e.g., sensor imaging theory, spatial correlation, and spectral characteristics of ground objects within RS images \cite{huang2023spectral}. For instance, a spectral
prompt tuning method is designed for feature enhancement using soft prompts \cite{kong2024joint}. Meanwhile, 
some hard prompts are also developed for designing specific prompt templates for different downstream tasks \cite{cao2023spectral,yang2024tmcfn}. Still, how to explore a more robust feature transmission and representation space to bridge various visual and linguistic information gaps remains necessary.}

The world model focuses on creating a latent interaction space for feature transmission across multimodal data. Especially, the fusion of VLMs and diffusion within the world model enhances its ability to bridge visual and linguistic gaps \cite{xu2022clip,clark2024text}, playing a significant role in robot navigation, game development, and autonomous driving \cite{wang2024driving,wu2023daydreamer,Zhao_2024_CVPR}. Recently, in the RS field, existing text-to-image models directly perform text and visual classification independently, lacking a unified feature expression space. To this end, the characteristic morphology of the world model is first introduced into the RS field, and a label-efficient remote sensing world model for multimodal data fusion (FusDreamer) is developed in this paper. On the one hand, the proposed FusDreamer can use a wide area of open-world knowledge to improve classification accuracies. On the other hand, joint training helps in aligning the feature spaces of text and multimodal data, thereby reducing the domain gap and improving the model's ability to generalize across different types of data. Compared with vision-centric-only models using specific visual features, the proposed FusDreamer uses a generative world model for text and visual information fusion and classification, which can enhance the model's generalization capabilities, making it more effective in handling diverse and complex RS tasks even with limited labeled samples. 
The main contributions are as follows:\par
\begin{itemize}
\item[1)] The first RS world model, namely FusDreamer, is developed, offering a unified representation container for multimodal data and promoting hierarchical understanding and transmission of features in a cohesive manner.\par
\item[2)] A new interactive latent diffusion paradigm is developed for multimodal feature generation, incorporating intrinsic visual information and physical knowledge during the feature reverse process through a combinatorial optimization strategy, thus promoting richer feature representation.\par
\item[3)] An open-world knowledge-guided consistency projection module integrates self-categorical and differentiated physical prompt representations. The pre-trained open-world knowledge facilitates domain-invariant learning for solving multimodal classification with limited labels.\par
\item[4)] Experimental results conducted on four multimodal datasets show the advantages of the FusDreamer, which can always achieve the highest performance when compared with state-of-the-art (SOTA) networks.\par
\end{itemize} 

The remainder of this paper is organized as follows. Section \ref{related work} describes the related works. Section \ref{proposed approach} presents the details of the proposed FusDreamer approach. In Section \ref{EXPERIMENTAL RESULTS AND DISCUSSION}, the experimental results and discussions are provided. Finally, conclusions are given in Section \ref{conclusions}.

\section{Related Works}
\label{related work}
{ \subsection{World Model}
World models are internal representations and simulations for specific environments, which can facilitate the simulation of complex scenarios without relying on real-world interactions and are crucial for tasks requiring high-level cognitive functions \cite{ha2018recurrent}.

 A world model is viewed as a generative model to facilitate the generation representation of features and enable perception, control, and prediction for downstream tasks \cite{gao2024enhance}. Typically, the world model uses latent variable models (such as the variational autoencoders (VAE) \cite{9136917}, etc.) to model the latent states of the environment.} Next, spatiotemporal dynamics modeling can be created by using recurrent neural networks (RNN) or transformer architectures \cite{han2022survey}, which can predict the evolution of a series of future states. Last, strategies and decisions related to specific goals are generated, where diffusion models serve as a mainstream structure of probabilistic generative models that have been widely used in the world model \cite{wang2023drivedreamer,zhao2024drivedreamer}. They progressively introduce noise to data and subsequently learn to reverse this process for the purpose of generating samples \cite{yang2023diffusion}. For instance, real-world-driven world models (DriveDreamer) \cite{wang2023drivedreamer} enable autonomous vehicles to comprehend complex driving scenarios by learning from real-world data. Building on this foundation, Llm-enhanced world models (DriveDreamer2)  \cite{zhao2024drivedreamer} elevate the world model to a higher level, enhancing the vehicle's ability to robustly understand, predict, and navigate intricate driving environments, ultimately contributing to safer and more reliable autonomous driving. {To sum up, the world model achieves huge success in autonomous driving due to the building of a latent modular interaction space for feature transmission across various multimodal data.
 
Inspired by the concept of the world model, this paper aims to provide a stable and unified representation space for simulating environments through generative models, compressing high-dimensional data into abstract representations, and integrating multimodal data to support downstream tasks. This approach introduces a novel paradigm for extending the world model to the remote sensing (RS) field.}

\begin{figure*}
	\centering
    \includegraphics[scale=0.95]{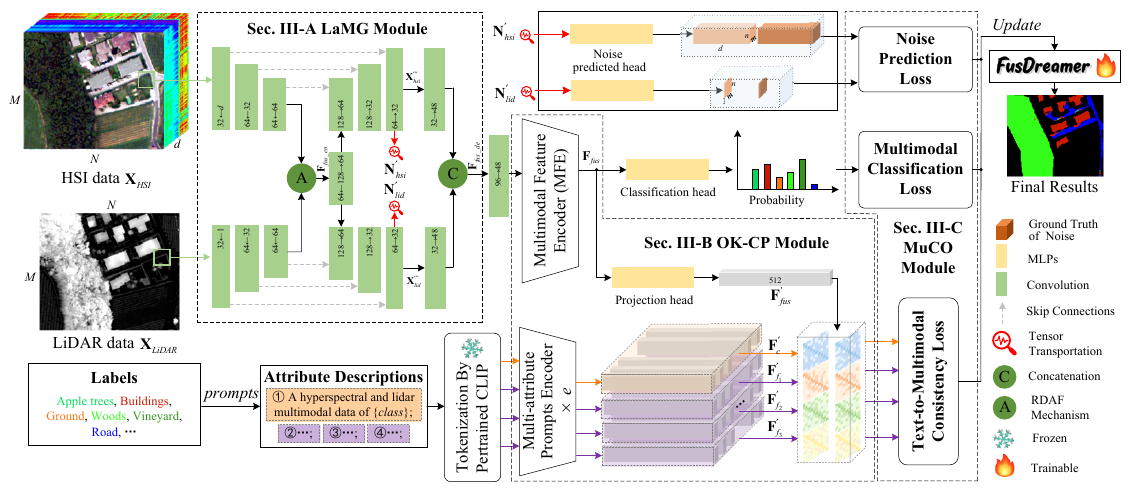}\par
\caption{The architecture of the proposed FusDreamer network, offering a unified representation container for latent feature generation and multimodal feature interaction. The LaMG module uses an interactive latent diffusion for generating learnable latent multimodal features. Then, the OK-CP module generates semantic-aware open-world prompts from physical knowledge attributes, and their pre-trained knowledge is well-suited for training multimodal data with few labeled samples. Finally, the MuCO module constrains the data generation by considering the open-world prompts.}
\label{fig:coarse_framework}
\end{figure*}



 \subsection{Pre-Training VLMs for Multimodal Data Fusion}
{To reduce the reliance on the need for labels, self-supervision, and knowledge transfer-based methods \cite{wang2022self,10483555} serve as an efficient direction to learn and generalize the model effectively, such as masked autoencoders (MAE) \cite{he2022masked} and a simple framework for masked image modeling (SimMIM) \cite{xie2022simmim}. In addition, researchers indicate that models pre-trained on large datasets, such as the ImageNet and other large-scale image-text pairs, can also be fine-tuned to improve the performance of specific RS tasks \cite{li2024vision}. Recently, VLMs have been popularly used in the RS field. 

There are two main classes of VLMs, \emph{i.e.}, single-branch and dual-branch structures. The single-branch structure adopts an early-stage fusion, jointly encoding image and text pairs to model visual and textual representations in a multi-layer cross-modal Transformer, such as a simple and performant baseline for vision and language (VisualBERT) \cite{li2019visualbert} and universal image-text representation (Uniter) \cite{chen2020uniter}. In contrast, the dual-branch structure employs a late-stage fusion, separately encoding image and text features and then using a dot product or multilayer perception to capture their interactions and obtain high-level representations of the modalities, such as a contrastive language-image pretraining (CLIP) \cite{radford2021learning} and a large-scale image and
noisy-text embedding (ALIGN) \cite{jia2021scaling}. For instance, Zhang \emph{et al.} utilize linguistic features for shared semantic space and align visuals with language through supervised contrastive learning, achieving strong domain adaptation and classification performance \cite{zhang2023language}. Cao \emph{et al.} propose a spectral-spatial-language fusion network (S2LFNet), which can broaden the semantic space using linguistic priori knowledge commonly shared between spectral features and spatial features and improve the data representation capabilities for multimodal data fusion tasks \cite{cao2023spectral}. Experiments indicate that the pre-training VLM is a new research direction for RS multimodal data fusion and classification.}

\section{The Proposed Approach}
\label{proposed approach}
{This paper develops a FusDreamer based on the concept of the world model, which can create a unified and integrated multimodal latent state representation space and help bridge multimodal information gaps, such as visual and linguistic information. The FusDreamer contains three main parts, \emph{i.e.}, the latent-spatial multimodal generation (LaMG) module in Sec. \ref{DAFM}, open-world knowledge-guided consistency projection (OK-CP) module in Sec. \ref{TMP}, and multitask combinatorial optimization (MuCO) module in Sec. \ref{CombinatorialOptimization}. 
Specifically, the LaMG module extracts inverse latent multimodal vision features through a latent diffusion process \cite{rombach2022high,zhao2023ddfm}. The OK-CP module generates semantic-aware open-world prompts based on physical knowledge attributes, enabling precise linguistic information description. Within the latent space of the world model, the visual and linguistic features interact and mutually constrain each other through an end-to-end multitask collaborative training framework  in the OK-CP module, synergistically enhancing the multimodal data fusion process. The entire framework is illustrated in Fig. \ref{fig:coarse_framework}. }

Let's denote an original hyperspectral image as $\mathbf{X}_{HSI}^D \in \mathbb{R}^{M \times N \times D}$ and an original LiDAR image as $\mathbf{X}_{LiDAR} \in \mathbb{R}^{M \times N \times 1}$, and let both images correspondingly cover the same surface area, where $M$ and $N$ refer to the two spatial dimensions of the images, and $D$ represents the number of spectral channels in the original HSI. Then, after a principal component analysis (PCA) operation, the dimensionality of the processed HSI becomes $\mathbf{X}_{HSI} \in \mathbb{R}^{M \times N \times d}$. For the HSI-LiDAR pair training set $\mathbf{X}$, the patched data are denoted as $\mathbf{X}_{hsi} \in \mathbb{R}^{m \times n \times d}$ and $\mathbf{X}_{lid} \in \mathbb{R}^{m \times n \times 1}$ with patch size $m\times n$. $\mathbf{Y}=\left\{y^{o}\right\}_{o=1}^{O}$ represents the corresponding groundtruth labels of the HSI-LiDAR pair, where $y^{o} \in\{1,2, \ldots, C\}$, and $O$ and $C$ denote the total number of the training samples and classes, respectively. 


\subsection{Latent-spatial Multimodal Generation Module}
\label{DAFM}
Typical CNN-based discriminative models grab multi-level information based on the structure of their feature extractors, which are minimally affected or restricted by semantic features from other domains. In contrast, generative models, such as the latent diffusion model \cite{rombach2022high}, have more flexibility in multimodal feature fusion and allow controlled data generation under specific conditions (such as considering semantic structure derived from physical knowledge attributes). Therefore, a latent diffusion-based generative model is more learnable and manageable, and it is used as a feature extractor to handle and integrate features from different modalities for HSI-LiDAR feature fusion generation. 

Specifically, the paper designs a LaMG module for RS feature generation. The forward diffusion and reverse generation processes follow the fixed learning strategy of DDPMs \cite{yang2023diffusion}, developing a unique adaptive and interactive multimodal data fusion strategy during the reverse generation process. Here, the forward diffusion process $q(\cdot^t|\cdot^{t-1})$ is mainly responsible for adding normally distributed noise data ${\bf{N}}_{hsi}$ and ${\bf{N}}_{lid}$ to the input multimodal data. 
\begin{subequations}
\begin{align}
q\left( {{\bf{X}}_{hsi}^t\left| {{\bf{X}}_{hsi}^{t - 1}} \right.} \right) = {\cal N}\left( {{\bf{X}}_{hsi}^t;\sqrt {1 - {\beta _t}} {\bf{X}}_{hsi}^{t - 1},{\beta _t}{{\bf{I}}_{hsi}}} \right)
, 
\label{EQ:1}\\
q\left( {{\bf{X}}_{lid}^t\left| {{\bf{X}}_{lid}^{t - 1}} \right.} \right) = {\cal N}\left( {{\bf{X}}_{lid}^t;\sqrt {1 - {\beta _t}} {\bf{X}}_{lid}^{t - 1},{\beta _t}{{\bf{I}}_{lid}}} \right)
,
\label{EQ:2}
\end{align}
\end{subequations}
where $t\in [0, T]$ and $T$ represent the total sampled number. $\cal N$ represents a normal distribution. $\sqrt {1- {\beta _t}}$ and ${\beta _t}$ represent the mean and variance.
Afterward, a multimodal reverse diffusion process $p(\cdot^{t-1}|\cdot^{t})$ is elaborated, which is responsible for optimizing and generating more discriminative features. 
\begin{equation}
\begin{aligned}
&p\left( {\left( {{\bf{X}}_{hsi}^{t - 1}\left| {{\bf{X}}_{lid}^{t - 1}} \right.} \right)\left| {\left( {{\bf{X}}_{hsi}^t\left| {{\bf{X}}_{lid}^t} \right.} \right)} \right.} \right) =\\& {\cal N}\left( {\left( {{\bf{X}}_{hsi}^{t - 1},{\bf{X}}_{lid}^{t - 1}} \right);{\mu _\theta }\left( {\left( {{\bf{X}}_{hsi}^{t - 1},{\bf{X}}_{lid}^{t - 1}} \right),t} \right),{\Sigma _\theta }\left( {\left( {{\bf{X}}_{hsi}^{t - 1},{\bf{X}}_{lid}^{t - 1}} \right),t} \right)} \right).
\end{aligned}
\end{equation}

The reverse diffusion process employs an encoder-decoder structure following an interactive U-Net, ensuring that shallow convolutions focus on texture features while deeper layers capture essential representations. More details are as below. 

\subsubsection{Reverse Diffusion Encoder (RDE)}
The encoder consists of $U$ residual convolution blocks, with a downsampling module inserted between every two residual blocks. {For the HSI encoder branch, the output of the $u$-th residual convolution block ${{\bf{R}}_u}\left( {{{\bf{X}}_{hsi}}} \right)$ can be formulated as:
\begin{equation}
{{\bf{R}}_{u}}\left( {{{\bf{X}}_{hsi}}} \right) = Res\left( {{{\bf{X}}_{hsi}}} \right) = \underbrace {{{\left\{ {\sigma \left( {BN\left( {Conv\left( {{{\bf{X}}_{hsi}}} \right)} \right)} \right)} \right\}}_i}}_{i = 2} + {{\bf{X}}_{hsi}},
\end{equation}
where $Res\left(\cdot\right)$, $Conv\left(\cdot\right)$, $BN\left(\cdot\right)$, and $\sigma\left(\cdot\right)$ represent the residual blocks, convolution layer, batch normalization layer, and non-linear activate layer, respectively.}

After obtaining ${{\bf{R}}_{u}}\left( {{{\bf{X}}_{hsi}}} \right)$, it undergoes a single downsampling step to enhance the receptive field, where the downsampling module ($Down(\cdot)$) consists of a max-pooling layer ($MaxPool(\cdot)$), a linear projection layer, and timestep embedding as below.
\begin{equation}
\begin{aligned}
{\bf{R}}_{u}^{{\prime}}\left( {{{\bf{X}}_{hsi}}} \right) &= Down\left( {{{\bf{R}}_{u}}\left( {{{\bf{X}}_{hsi}}} \right)} \right)\\
&= Conv\left( {MaxPool\left( {{{\bf{R}}_{u}}\left( {{{\bf{X}}_{hsi}}} \right)} \right)} \right) \\
&\ \ \ + TimeEmbedding\left( t \right)
\end{aligned}.
\end{equation}

To effectively capture details and edge information from the high-resolution feature tensor and to help the model better understand the complex structure and subtle differences of the input data, no downsampling steps are applied after the final layer of residual blocks.

For more details, the number of residual convolution blocks, $U$, is set to three, with the number of convolution filters in these blocks set to 32, 64, and 64, respectively. The kernel size and stride of the convolution operation are set to 3 and 1, respectively. The output of the RDE can be expressed as:
\begin{equation}
{\bf{X}}_{hsi}^{{\prime}} = {{\bf{R}}_{u = 2}}\left( {\underbrace {{\bf{R}}_u^\prime \left( {{\bf{R}}_u^{}\left( {{{\bf{X}}_{hsi}}} \right)} \right)}_{u = \left[ {0,1} \right]}} \right),
\end{equation}
where ${\bf{X}}_{hsi}^{{\prime}} \in \mathbb{R}^{\frac{M}{4} \times \frac{N}{4} \times 64}$. Similarly, the output of LiDAR encoder branch ${\bf{X}}_{lid}^{'}$ can also be obtained with shape of ${\bf{X}}_{lid}^{{\prime}} \in \mathbb{R}^{\frac{M}{4} \times \frac{N}{4} \times 64}$.

\subsubsection{Reverse Diffusion Adaptive Fusion (RDAF)}
Reverse adaptive fusion is performed on the encoded features through the designed adaptive weighted fusion strategy. Adaptive weighting with learnable parameters enables the model to automatically adjust the weight of each feature tensor, i.e., ${\rm{M}_{hsi}}$ and $ {\rm{M}_{lid}}$, according to the data characteristics during training, thereby achieving more flexible and effective feature fusion. The formulation can be expressed by:
\begin{equation}
{{\rm{M}}_{hsi}},{{\rm{M}}_{lid}} = Softmax\left( {Conv\left( {Concat\left( {{\bf{X}}_{hsi}^{'},{\bf{X}}_{lid}^{{\prime}}} \right)} \right)} \right),
\end{equation}
where $Concat(\cdot)$ represents the concatenation operation. The learnable weight can dynamically adjust the contribution of diffusion features and highlight critical features between multimodal data as follows:
\begin{equation}
{{\bf{F}}_{fus\_en}} = Conv\left(Concat\left( {{\bf{X}}_{hsi}^{{\prime}} \otimes {{\rm{M}}_{hsi}},{\bf{X}}_{lid}^{{\prime}} \otimes {{\rm{M}}_{lid}}} \right)\right),
\end{equation}
where ${{\bf{F}}_{fus\_en}} \in \mathbb{R}^{\frac{M}{4} \times \frac{N}{4} \times 128}$.

\subsubsection{Reverse Diffusion Decoder (RDD)} 
The decoder uses a symmetrical network structure with the encoder and performs feature restoration and propagation following three residual blocks of the decoder. At the same time, an upsampling module is inserted between every two residual blocks to gradually restore the spatial size of the feature map during the decoding stage, therefore, high-level semantic information can be propagated into a higher-resolution feature map.

{Specifically, the fused feature ${{\bf{F}}_{fus\_en}}$ is first mapped to the data space of HSI and LiDAR, respectively through two independent linear mapping layers, which can be expressed as ${{\bf{X}}_{hsi}^{{\prime}{\prime}}} \in \mathbb{R}^{\frac{M}{4} \times \frac{N}{4} \times 64}$ and ${{\bf{X}}_{lid}^{{\prime}{\prime}}}  \in \mathbb{R}^{\frac{M}{4} \times \frac{N}{4} \times 64}$.
\begin{equation}
{{\bf{R}}_{v}}\left( {{{\bf{X}}_{hsi}^{{\prime}{\prime}}}} \right) = Res\left( {{\bf{X}}_{hsi}^{{\prime}{\prime}}} \right) = \underbrace {{{\left\{ {\sigma \left( {BN\left( {Conv\left( {{\bf{X}}_{hsi}^{{\prime}{\prime}}} \right)} \right)} \right)} \right\}}_i}}_{i = 2} + {{\bf{X}}_{hsi}^{{\prime}{\prime}}},
\end{equation}
where $v$ represents the $v$-th residual block in the reverse diffusion decoder.} 

Once the decoding diffusion feature reverses, an upsampling module ($Up(\cdot)$) is used to ensure consistency with the feature tensor of the symmetrical $u$-th encoding layer. At the same time, Combined with skip connections, the high-resolution features retained in the encoder are fused with the upsampled features in the decoder, enhancing feature representation and further improving the model's accuracy and performance as shown below.
\begin{equation}
\begin{aligned}
{{\bf{R}}_{v}^{{\prime}}}\left( {{{\bf{X}}_{hsi}^{{\prime}{\prime}}}} \right) =& Up\left({{\bf{R}}_{v}}\left( {{{\bf{X}}_{hsi}^{{\prime}{\prime}}}} \right)\right)\\
=& Conv\left(Concat\left({{\bf{R}}_{u}}\left( {{{\bf{X}}_{hsi}}} \right),Upsample({{{\bf{X}}_{hsi}^{{\prime}{\prime}}}})\right)\right) \\
& + TimeEmbedding\left( t \right).
\end{aligned}
\end{equation}

Here, the number of residual convolution blocks $V$ is set to three. The number of convolution filters in three stages adopts 64, 32, and 32, respectively. Similar to the encoder, the output of the decoder can be expressed as:
\begin{subequations}
\begin{align}
{\bf{X}}_{hsi}^{{\prime}{\prime}{\prime}} = {{\bf{R}}_{v = 2}}\left( {\underbrace {{\bf{R}}_v^\prime \left( {{\bf{R}}_v^{}\left( {{\bf{X}}_{hsi}^{\prime \prime }} \right)} \right)}_{v = \left[ {0,1} \right]}} \right),
\label{EQ:3}\\
{\bf{X}}_{lid}^{{\prime}{\prime}{\prime}} = {{\bf{R}}_{v = 2}}\left( {\underbrace {{\bf{R}}_v^\prime \left( {{\bf{R}}_v^{}\left( {{\bf{X}}_{lid}^{\prime \prime }} \right)} \right)}_{v = \left[ {0,1} \right]}} \right),
\label{EQ:4}
\end{align}
\end{subequations}
where ${\bf{X}}_{hsi}^{{\prime}{\prime}{\prime}} \in \mathbb{R}^{M \times N \times 32}$ and ${\bf{X}}_{lid}^{{\prime}{\prime}{\prime}} \in \mathbb{R}^{M \times N \times 32}$. Afterward, the output of decoder ${{\bf F}_{fus\_de}}$ can be obtained by:
\begin{equation}
{{\bf F}_{fus\_de}} = Concate\left(Conv\left({\bf{X}}_{hsi}^{{\prime}{\prime}{\prime}}\right),Conv\left({\bf{X}}_{lid}^{{\prime}{\prime}{\prime}}\right)\right),
\end{equation}
where ${{\bf F}_{fus\_de}}  \in \mathbb{R}^{m \times n \times 96}$.

To optimize the latent diffusion module, it is necessary to estimate the error between the predicted noise information and the added noise data. The predicted noise information (${\bf{N}}_{hsi}^{{\prime}}$ and ${\bf{N}}_{lid}^{{\prime}}$) can be separated from the output of the decoder. Specifically, by post-processing the decoder's output, the noise components can be effectively extracted: 
\begin{equation}
{\bf{N}}_{hsi}^{{\prime}} = Conv\left( {{\bf{X}}_{hsi}^{{\prime}{\prime}{\prime}}} \right), {\bf{N}}_{lid}^{{\prime}} = Conv\left( {{\bf{X}}_{lid}^{{\prime}{\prime}{\prime}}} \right).
\end{equation}

The specific loss function to optimize the diffusion module is described in detail in Sec. \ref{CombinatorialOptimization}, directly affecting the reverse process of latent diffusion generation.

\subsection{Open-World Knowledge-guided Consistency Projection Module}
\label{TMP}
Vision-centric models often struggle to train effectively with limited labeled samples. Research suggests that incorporating open-world prompt information can enhance feature representation, as it provides a deep understanding of objects' physical attributes and their relationships. Therefore, this paper integrates the pre-trained open-world knowledge into the world model, following a CLIP structure for prompt-multimodality generation. More details are as below.
\subsubsection{Multimodal Feature Encoder (MFE)}
\label{MFE}
To fully preserve the attributes of the multimodal fused feature tensor in three-dimensional space, 3D residual convolution layers are employed to encode the fused features at a deeper level. These 3D residual convolution layers not only capture richer spatial information but also effectively extract cross-modal joint features. Additionally, skip connections are incorporated into the deep encoding module, helping transmit more detailed information across multiple network layers and thereby improving the overall performance of the model.
\begin{equation}
{{\bf{F}}_{fus}} = \sigma \left( {3DConv\left( {3DConv\left( {{{\bf{F}}_{fus\_de}}} \right)} \right) + 3DConv\left( {{{\bf{F}}_{fus\_de}}} \right)} \right).
\end{equation}

\begin{figure}
    \centering
    \includegraphics[scale=0.99]{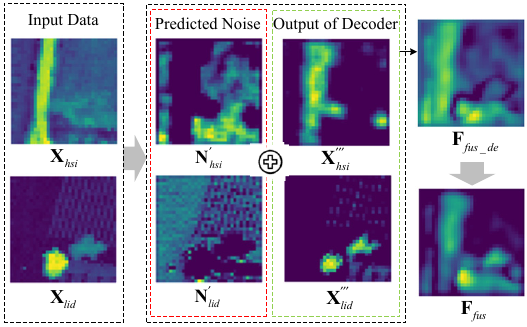}\par
    \caption{An illustration of multimodal data feature learning process.}
    \label{fig:feature}
\end{figure}
Fig. \ref{fig:feature} illustrates the feature extraction process for multimodal data. As depicted in the figure, the fused decoding feature ${{\bf{F}}_{fus\_de}}$ derived from the generative LaMG module can effectively integrate key features from the multimodal input data. However, it also retains interference information from the original data. In contrast, the fused feature ${{\bf{F}}_{fus}}$ obtained after passing through the multimodal feature encoder significantly suppresses background noise, thereby enhancing the consistency projection between the prompts and the multimodal data.

The classification result of each pixel $\mathord{\buildrel{\lower3pt\hbox{$\scriptscriptstyle\frown$}} \over y}$ can be obtained by using the multimodal fusion features. 
\begin{equation}
\mathord{\buildrel{\lower3pt\hbox{$\scriptscriptstyle\frown$}} \over y}  = Soft\max \left( {FC\left( {{{\bf{F}}_{fus}}} \right)} \right).
\end{equation}

To optimize the multimodal fusion module by using the knowledge information of the world dimension, it is also necessary to map the fusion features ${{{\bf{F}}_{fus}}}$ to the spatial dimension ${{\bf{F}}_{fus}^{\prime}}$ and perform structural mapping with the open-world encoding features. This mapping allows for better capture and utilization of interactive information between different modalities.
\begin{equation}
{{\bf{F}}_{fus}^{\prime}} = FC\left( {{{\bf{F}}_{fus}}} \right).
\end{equation}

\begin{table}
  \centering
  \scriptsize
  \caption{Examples of Prompts for the Trento Dataset.}
    \begin{tabular}{cp{27.5em}}
    \toprule
    Classes & \multicolumn{1}{c}{Prompts} \\
    \midrule
    \multirow{4}[0]{*}{Apple trees} &    \textcolor[rgb]{1, 0.5, 0}{A hyperspectral and lidar multimodal data of apple trees}\\
                                    &  \textcolor[rgb]{0.44, 0.19, 0.63}{The apple trees appear khaki and green} \\
                                    &  \textcolor[rgb]{0.44, 0.19, 0.63}{The height of apple trees is lower than that of vineyard} \\
                                    &  \textcolor[rgb]{0.44, 0.19, 0.63}{There is ground in the middle of the gap between apple trees} \\
    \midrule
    \multirow{4}[0]{*}{Buildings} &   \textcolor[rgb]{1, 0.5, 0}{A hyperspectral and lidar multimodal data of buildings}\\
                                  & \textcolor[rgb]{0.44, 0.19, 0.63}{The buildings are more distant from the ground than from the road} \\
                                  & \textcolor[rgb]{0.44, 0.19, 0.63}{The buildings are well spaced} \\
                                  & \textcolor[rgb]{0.44, 0.19, 0.63}{The height of buildings is close to the woods} \\
    \midrule
    \multirow{4}[0]{*}{Ground} &   \textcolor[rgb]{1, 0.5, 0}{A hyperspectral and lidar multimodal data of ground}\\
                               & \textcolor[rgb]{0.44, 0.19, 0.63}{The ground occurs next to apple trees} \\
                               & \textcolor[rgb]{0.44, 0.19, 0.63}{The length of the ground is shorter than the length of the road} \\
                               & \textcolor[rgb]{0.44, 0.19, 0.63}{The ground has the lowest height} \\
    \midrule
    \multirow{4}[3]{*}{Woods} &    \textcolor[rgb]{1, 0.5, 0}{A hyperspectral and lidar multimodal data of woods}\\
                              &  \textcolor[rgb]{0.44, 0.19, 0.63}{The woods appear as small circles} \\
                              &  \textcolor[rgb]{0.44, 0.19, 0.63}{The woods are dark green} \\
                              &  \textcolor[rgb]{0.44, 0.19, 0.63}{The height of the woods has a clear fluctuation trend in the local area} \\
    \midrule
    \multirow{4}[0]{*}{Vineyard}  &    \textcolor[rgb]{1, 0.5, 0}{A hyperspectral and lidar multimodal data of vineyard}\\
                                  &  \textcolor[rgb]{0.44, 0.19, 0.63}{Vineyard is a regular rectangle or square} \\
                                  &  \textcolor[rgb]{0.44, 0.19, 0.63}{Vineyard and apple trees are far away} \\
                                  &  \textcolor[rgb]{0.44, 0.19, 0.63}{The spectral values of the vineyard and apple trees are close} \\
    \midrule
    \multirow{4}[0]{*}{Road}  & \textcolor[rgb]{1, 0.5, 0}{A hyperspectral and lidar multimodal data of road}\\
                              & \textcolor[rgb]{0.44, 0.19, 0.63}{Trees grew along the road} \\
                              & \textcolor[rgb]{0.44, 0.19, 0.63}{The buildings are next to road} \\
                              & \textcolor[rgb]{0.44, 0.19, 0.63}{The road appears as an elongated strip shape} \\
    \bottomrule
    \end{tabular}
  \label{tab:prompttr}
\end{table}

\subsubsection{Multi-attribute Prompts Encoder (MPE)}
{
The MPE incorporates a broader range of descriptors—color, shape/height, and inter-class relationships—across four distinct prompt types. Leveraging prior knowledge of land cover classes within the remote sensing (RS) scene, both self-categorical and differentiated physical descriptions are carefully crafted for each class. As illustrated in Table \ref{tab:prompttr}, ``\textcolor[rgb]{1, 0.5, 0}{A hyperspectral and lidar multimodal data of $<class\ name>$}'' is employed as a template to generate self-categorical prompt descriptions ${T}_c$ for each class using a cloze format. For differentiated physical descriptions $\left[{T}_{d_1}, {T}_{d_2}, {T}_{d_3}\right]$, prior knowledge is used to manually describe attributes such as color, shape, distribution, and adjacency relationships; examples include [``\textcolor[rgb]{0.44, 0.19, 0.63}{The apple trees appear khaki and green}'', ``\textcolor[rgb]{0.44, 0.19, 0.63}{The buildings are next to road}'', and ``\textcolor[rgb]{0.44, 0.19, 0.63}{Vineyard and apple trees are far away}'']. Each land cover class is assigned three differentiated physical descriptions, marked by the purple color in Table \ref{tab:prompttr}.}

First, these prompts need to be tokenized by the pre-trained language model adjusted following Ref. \cite{radford2019language}. It begins with a foundational model consisting of 33M parameters, organized into 3 layers, with a width of 512 and 8 attention heads. Similar to CLIP, the transformer uses lowercase byte pair encoding (BPE) for text representation, with a vocabulary size of 49,152. To ensure computational efficiency, the sequence length is capped at 77. These linguistic features are then layer-normalized and linearly projected into the semantic space.
\begin{subequations}
\begin{align}
{{\bf{F}}_c} &= CLIP\left( {{T_c}} \right),
\label{EQ:5}\\
{{\bf{F}}_{{d_1}}},{{\bf{F}}_{{d_2}}},{{\bf{F}}_{{d_3}}} &= CLIP\left( {{T_{{d_1}}}} \right),CLIP\left( {{T_{{d_2}}}} \right),CLIP\left( {{T_{{d_3}}}} \right),
\label{EQ:6}
\end{align}
\end{subequations}
where ${{\bf{F}}_c} \in \mathbb{R}^{1\times77}$, the real number units for ${{\bf{F}}_{{d_1}}} $, ${{\bf{F}}_{{d_2}}}$, and ${{\bf{F}}_{{d_3}}}$ are consistent with those of ${{\bf{F}}_c}$. Afterward, a weight-sharing multi-branch prompt encoding module is employed to map specific self-categorical and differentiated physical descriptions into the open-world dimension, obtaining their representational features in the world domain. 

Then, tokenized prompts are fed into the post-processing stage of the MPE module, which mainly consists of multiple cascaded transformers, where the number of transformer layers, $e$, is a hyperparameter to be discussed in Sec. \ref{transformernumber}.
\begin{equation}
\begin{aligned}
{\bf{F}^{\prime}} = \underbrace{Transformer\left( {\bf{F}} \right)}_{\times e},\ \
\end{aligned}
\end{equation}
where ${\bf{F}}$ represents an example input of the prompt encoding module. Thus, the representation of self-categorical and differentiated physical texts in the world dimension can be expressed as ${{\bf{F}}_c^{\prime}}$, ${{\bf{F}}_{{d_1}}^{\prime}}$, ${{\bf{F}}_{{d_2}}^{\prime}}$, and ${{\bf{F}}_{{d_3}}^{\prime}}$, respectively.




\subsubsection{Prompts-Multimodality Alignment}To map features of prompt and multimodality into a common representation space, it is necessary to compute the similarity among them, thereby optimizing the multimodal feature fusion framework through contrastive learning. Here, ${\bf F}_{fus}^{\prime}$ represents the high-dimensional feature representations of the multimodal data, and ${\bf F}_{c}^{\prime}$ represents the high-dimensional feature representations of the multimodal and prompt information, respectively. By calculating the cosine similarity $Cor_{m 2 c}$ and $Cor_{c2m}$ between multimodal feature ${{\bf{F}}_{fus}^{\prime}}$ and text features of self-categorical prompt ${{\bf{F}}_c^{\prime}}$, the degree of match between prompt-multimodality pairs can be quantified, which is used for subsequent contrastive learning and optimization. Specifically, it can be obtained by: 
\begin{equation}
Co{r_{m 2 c}} = {\bf{F}}_c^\prime  \otimes {\left( {{\bf{F}}_{fus}^\prime } \right)^T},Co{r_{c 2 m}} = {\bf{F}}_{fus}^\prime  \otimes {\left( {{\bf{F}}_c^\prime } \right)^T},
\label{EQ:7}
\end{equation}
where $Cor_{m2c}$ is from the prompt perspective, aiding in representing prompt features within the context of fused multimodal features. Comparatively, $Cor_{c2m}$ is from the multimodal data perspective to capture the prompt information.

Similarly, the cosine similarity between prompt-multimodality features of differentiated physical prompts $\left(\left(Cor_{m2{d_1}}, Cor_{{d_1}2m}\right), \left(Cor_{m2{d_2}}, Cor_{{d_2}2m}\right), \left(Cor_{m2{d_3}}, Cor_{{d_3}2m}\right)\right)$ can be obtained similarly.

\subsection{Multitask Combinatorial Optimization Module}
\label{CombinatorialOptimization}
This paper adopts a MuCO module to leverage large-scale prior information from the open world to optimize the multimodal feature fusion module. This strategy can directly constrain the data generation of the reverse process of the LaMG module by considering the open-world prompts derived from physical knowledge, synergistically optimizing the diffusion noise prediction module, the multimodal classification module, and the prompts-multimodality feature projection module. 

\subsubsection{Noise Prediction Loss} By incorporating large-scale prior information, the model's ability to understand and process different modalities can be enhanced, thereby effectively predicting and handling noise within the framework of multitask learning. 
\begin{equation}
\mathscr{Loss}_{N} ={\frac{1}{2}} \left(MSE\left( {{\bf{N}}_{hsi} ,{\bf{N}}_{hsi}^\prime } \right)+ MSE\left( {{\bf{N}}_{lid} ,{\bf{N}}_{lid}^\prime } \right) \right),
\end{equation}
where $MSE(\cdot)$ represents the mean squared error function.

\subsubsection{Multimodal Classification Loss} Additionally, optimizing the multimodal fusion and classification module can improve the model's accuracy and robustness when dealing with complex multimodal data.
\begin{equation}
\mathscr{Loss}_C = CrEn\left(\mathord{\buildrel{\lower3pt\hbox{$\scriptscriptstyle\frown$}} \over y},y\right),
\end{equation}
where $CrEn(\cdot)$ represents the cross entropy loss function.
 
\subsubsection{Prompts-Multimodality Consistency Loss} Aligning prompt features with multimodal features can further enhance the model's feature representation capabilities and cross-modal retrieval performance. In the proposed method, self-categorical prompts are used to guide the model's generalization performance in complex scenarios under the guidance of prior information from the world model dimension. On the other hand, differentiated physical prompts help the model achieve high-precision and high-robustness feature fusion under specific constraints. Therefore, this module will separately handle these two sets of loss, i.e. $\mathscr{Loss}_{mc}$ and $\mathscr{Loss}_{md}$, to optimize the model's performance at different levels.
\begin{equation}
\mathscr{Loss}_{mc} = {\frac{1}{2}}\left(CrEn\left(Co{r_{m 2 c}},y\right) + CrEn\left(Co{r_{c 2 m}},y\right)\right),
\end{equation}
\begin{equation}
\begin{aligned}
\mathscr{Loss}_{md}= {\frac{1}{3}}&({\frac{1}{2}}\left(CrEn\left(Co{r_{m 2 {d_1}}},y\right) + CrEn\left(Co{r_{{d_1} 2m }},y\right)\right)\\
                                  & + {\frac{1}{2}}\left(CrEn\left(Co{r_{m 2 {d_2}}},y\right) + CrEn\left(Co{r_{{d_2}2m}},y\right)\right)\\
                                  & + {\frac{1}{2}}\left(CrEn\left(Co{r_{m 2 {d_3}}},y\right) + CrEn\left(Co{r_{{d_3}2m}},y\right)\right)) 
\end{aligned}.
\end{equation}

The prompts-to-multimodal feature consistency loss can be leveraged as:
\begin{equation}
\mathscr{Loss}_{M} = \alpha \mathscr{Loss}_{mc}+\left(1-\alpha\right) \mathscr{Loss}_{md},
\end{equation}
where $\alpha$ is a hyperparameter, with details in Sec. \ref{Implementdetail}.

Once the aforementioned loss information is accurately aggregated, the final loss $\mathscr{Loss}$ can be formulated as:
\begin{equation}
\mathscr{Loss} = \lambda_1  \cdot \mathscr{Loss}_C + \lambda_2  \cdot \mathscr{Loss}_N + \lambda_3  \cdot \mathscr{Loss}_{M} ,
\end{equation}
where $\sum\limits_{i = 1}^3 {\lambda_i} = 1$ represents the different weights assigned to the three loss function components. The impact of these weights on model optimization will be analyzed in detail in Sec. \ref{loss}.

\begin{table*}
  \centering
 { \caption{Quantitative Performance (in \%) of Different Classification Networks on the Houston 2013 Dataset. The Number in Parentheses Indicates the Standard Variance for the 20 Repeated Experiments, and the Best Classification Results are Shown in Bold.}
  \footnotesize
    \begin{tabular}{cccccccccc}
    \toprule
    \multicolumn{1}{c}{No.} & Class(Train/Test)  & MAHiDFNet &  AM$^3$Net & NNCNet & CALC & MBFormer & DSHFNet & FusDreamer \\
        \midrule
    1     & \hspace{0.8 em} Grass-healthy (20/1231)   & 88.42(5.59) & 61.27(6.91) & \textbf{92.40(1.23)} & 86.58(6.20) & 83.87(4.98) & 91.08(2.24) & 91.71(4.17) \\
    2     & \hspace{0.6 em} Grass-stressed (19/1235)  & 90.98(4.79) & 46.01(5.29) & \textbf{94.80(2.19)} & 81.65(17.4) & 84.87(5.36) & 91.36(1.51) & 90.20(7.55) \\
    3     & Grass-synthetic (19/679)                  & 90.70(2.07) & 85.05(7.71) & \textbf{99.85(0.00)} & 99.68(0.40) & 95.62(5.04) & 90.36(0.90) & 97.41(3.38) \\
    4     & \hspace{4.62 em} Tree (19/1225)           & 94.52(4.05) & 51.02(4.30) & \textbf{98.50(1.43)} & 96.34(2.59) & 70.01(11.7) & 92.51(0.30) & 92.14(8.18) \\
    5     & \hspace{4.7 em} Soil (19/1223)            & 97.58(1.50) & 87.09(1.22) & 93.18(5.53) & \textbf{99.51(0.73)} & 98.57(1.66) & 90.35(0.31) & 94.68(7.99) \\
    6     & \hspace{3.4 em} Water (18/307)            & \textbf{95.56(4.62)} & 95.42(2.64) & 89.90(2.02) & 84.63(3.15) & 90.57(4.32) & 90.61(1.31) & 89.59(5.00) \\
    7     & \hspace{1.8 em} Residential (19/1249)     & 89.51(4.75) & 72.44(3.82) & 87.12(4.06) & \textbf{94.74(6.54)} & 73.53(10.4) & 80.00(2.01) & 81.53(7.29) \\
    8     & \hspace{1.4 em} Commercial (19/1225)      & \textbf{88.98(7.39)} & 71.97(6.05) & 70.86(13.2) & 82.84(9.79) & 82.81(5.17) & 80.52(1.30) & 88.59(10.4) \\
    9     & \hspace{4.3 em} Road (19/1233)            & \textbf{85.66(7.50)} & 34.19(3.55) & 52.20(23.9) & 56.27(9.59) & 67.86(6.68) & 66.96(2.81) & 76.59(5.65) \\
    10    & \hspace{2.8 em} Highway (19/1208)         & 68.45(7.08) & 85.59(5.93) & \textbf{93.38(4.17)} & 79.32(21.2) & 89.10(10.7) & 83.71(1.12) & 84.76(11.2) \\
    11    & \hspace{3.2 em} Railway (18/1217)         & 86.48(1.32) & 85.86(4.46) & 94.74(2.30) & \textbf{94.81(6.51)} & 92.05(3.65) & 91.62(1.14) & 94.48(5.70) \\
    12    & \hspace{0.86 em} Parking Lot-1 (19/1214)  & 70.30(10.8) & 35.87(2.46) & 85.77(12.5) & 82.65(8.50) & 84.14(2.49) & 85.12(3.31) & \textbf{86.54(6.16)} \\
    13    & \hspace{0.35 em} Parking Lot-2 (18/451)   & 94.53(2.92) & 73.75(7.04) & 55.34(34.8) & \textbf{95.43(0.30)} & 92.94(6.91) & 88.54(0.21) & 85.94(12.6) \\
    14    & \hspace{0.72 em} Tennis Court (18/410)    & 93.89(3.22) & 98.36(1.79) & 86.10(13.3) & 89.66(21.9) & \textbf{100.0(0.00)} & 93.36(0.31) & 99.18(0.94) \\
    15    & \hspace{0. em} Running Track (19/641)     & 96.96(1.79) & 74.51(7.11) & \textbf{99.34(1.22)} & 95.76(4.20) & 97.83(1.96) & 95.34(0.10) & 98.92(0.94) \\
        \midrule
    \multicolumn{2}{c}{OA} & 86.94(1.32) & 66.22(1.21) & 86.58(3.62) & 86.97(2.38) & 84.67(0.86) & 87.15(0.41) &    \textbf{89.24(2.06)} \\
    \multicolumn{2}{c}{AA} & \textbf{88.83(1.18)} & 70.56(1.17) & 85.48(3.93) & 85.92(2.57) & 83.42(0.93) & 87.43(0.38)   & 88.35(2.23) \\
    \multicolumn{2}{c}{Kappa} & 85.88(1.42) & 63.51(1.30) & 86.23(4.55) & 87.99(2.61) & 86.92(0.61) & 87.95(0.43) & \textbf{90.15(1.94)} \\
        \hdashline
       \multicolumn{2}{c}{Training Time (s)}     &  \multicolumn{1}{r}{184.01}  & \multicolumn{1}{r}{34.13}  & \multicolumn{1}{r}{323.64}  & \multicolumn{1}{r}{871.94}  & \multicolumn{1}{r}{31.78}  & \multicolumn{1}{r}{665.78} & \multicolumn{1}{r}{75.27}  \\
      \multicolumn{2}{c}{Test Time (s)}          & \multicolumn{1}{r}{29.78}  & \multicolumn{1}{r}{3.51}  & \multicolumn{1}{r}{21.33}  & \multicolumn{1}{r}{61.30}  & \multicolumn{1}{r}{27.49}  & \multicolumn{1}{r}{22.15} & \multicolumn{1}{r}{30.05}  \\
    \bottomrule
    \end{tabular}%
  \label{tab:Houston2013small}}
\end{table*}%

\section{Experimental results and discussion}
\label{EXPERIMENTAL RESULTS AND DISCUSSION}

\subsection{Experiment Setup}

\subsubsection{Houston 2013 Dataset} It was collected in 2012 using airborne sensors over the University of Houston campus and surrounding urban areas. The Houston 2013 dataset includes both HSI and LiDAR-derived DSM, each with a data size of 349 $\times$ 1905 pixels and a spatial resolution of 2.5 $m$. The HSI consists of 144 spectral bands covering the wavelength range from 0.38 to 1.05 $\mu m$. This dataset includes a total of 15,029 ground-truth samples, with 15 distinct categories being analyzed.

\subsubsection{Houston 2018 Dataset} The coverage area of the Houston 2018 dataset is the same as the University of Houston campus. There are 48 spectral bands contained in the HSI data, ranging from 0.38 to 1.05 $\mu m$. The spatial size of this dataset is 601 $\times$ 2384 pixels, and the corresponding spatial resolution is 1 $m$ ground sampling distance. This dataset encompasses 20 distinct ground objects.

\subsubsection{MUUFL Dataset} It was taken at the University of Southern Mississippi Gulfport Campus, Mississippi, USA, in 2010. The HSI is composed of 325 $\times$ 220 pixels with 64 available spectral channels ranging from 0.38 to 1.05 $\mu$m, and the LiDAR modality consists of the elevation of two raster data.
This dataset includes a total of 53,687 ground-truth samples, with 11 distinct category labels being examined.

\subsubsection{Trento Dataset} It was captured in a rural area in southern Trento, Italy, and includes one HSI in 63 spectral bands and one LiDAR data. Both datasets have a size of 166 $\times$ 600 pixels with a spatial resolution of 1 $m$. The HSI covers a wavelength range of HSI from 0.42 to 0.99 $\mu m$. This dataset comprises a total of 30,214 ground-truth samples, with 6 distinct categories being analyzed.

\subsubsection{Performance Indices} To effectively evaluate the multimodal fusion performance of various networks, there are four common objective Indices here: class accuracy (CA), overall accuracy (OA), average accuracy (AA), and the Kappa coefficient (Kappa). Specifically, the CA value represents the percentage of correctly classified pixels in each class, while the OA determines the percentage of all test pixels that are correctly classified. The AA value records the mean of all class accuracies. The Kappa statistic is a multivariate statistical method that determines classification accuracy by considering the uncertainty factors in the classification process.

\begin{table*}
  \centering
 { \caption{Quantitative Performance (in \%) of Different Classification Networks on the Houston 2018 Dataset. The Number in Parentheses Indicates the Standard Variance for the 20 Repeated Experiments, and the Best Classification Results are Shown in Bold.}
  \footnotesize
  \setlength{\tabcolsep}{2 mm}
    \begin{tabular}{ccccccccc}
    \toprule
    No.   & Class(Train/Test)                                 & MAHiDFNet &  AM$^3$Net & NNCNet & CALC & MBFormer & DSHFNet & FusDreamer \\
    \midrule
    1     & \hspace{3.3 em} Healthy grass (50/9749)           & 62.28(13.9) & 51.23(4.77) & 79.79(5.81) & \textbf{93.95(1.87)} & 85.57(8.04) & 87.77(4.01) & 88.08(3.59) \\
    2     & \hspace{3.6 em} Stressed grass (50/32452)         & 84.48(3.90) & 48.89(8.64) & 68.28(5.22) & 53.50(4.50) & 68.83(9.53) & \textbf{85.53(4.18)} & 78.68(4.46) \\
    3     & \hspace{2.4 em} Artificial turf (7/677)           & 88.95(6.45) & -     & 91.52(6.29) & -     & \textbf{99.63(0.46)} & 43.38(44.4) & 99.10(0.73) \\
    4     & \hspace{3.0 em} Evergreen trees (50/13545)        & 83.94(5.41) & 75.29(6.09) & 91.25(2.57) & 85.55(0.79) & 96.33(1.41) & 90.54(1.12) & \textbf{97.42(1.07)} \\
    5     & \hspace{2.4 em} Diciduoud trees (50/4971)         & 29.10(5.44) & 67.76(4.12) & 81.98(3.81) & 21.99(2.92) & 90.05(8.04) & 80.31(6.91) & \textbf{91.80(1.60)} \\
    6     & \hspace{4.6 em}  Bare earth (45/4471)             & 44.12(9.80) & 96.65(1.76) & 91.49(2.38) & 94.51(1.07) & 99.09(1.68) & 91.51(0.15) & \textbf{99.15(0.33)} \\
    7     & \hspace{5.5 em} Water (3/263)                     &   78.02(17.3) & -     & 88.74(6.89) & -     & 36.39(22.5) & -     & \textbf{95.19(6.91)} \\
    8     & \hspace{0.9 em}  Residential buildings (50/39722) & 64.13(3.12) & 90.87(2.88) & \textbf{93.76(2.04)} & 84.47(0.52) & 84.46(4.77) & 84.19(2.52) & 81.16(3.32) \\
    9     & Non-residential buildings (50/223702)             & \textbf{98.46(0.75)} & 80.66(5.82) & 87.57(2.41) & 74.25(0.88) & 83.12(3.72) & 70.51(2.49) & 84.92(1.52) \\
    10    & \hspace{6.9 em} Roads (50/45816)                  & \textbf{58.58(6.19)} & 35.47(9.22) & 39.92(4.42) & -     & 43.09(8.29) & 15.95(16.1) & 48.14(4.74) \\
    11    & \hspace{5.4 em} Sidewalks (50/33979)              & \textbf{60.09(3.88)} & 36.82(9.89) & 37.74(6.82) & 3.67(0.48) & 38.64(9.33) & 28.91(5.17) & 49.56(5.74) \\
    12    & \hspace{4.4 em} Crosswalks (15/1503)              & 6.87(1.82) & -     & \textbf{38.14(6.58)} & -     & 18.78(2.23) & -     & 24.72(4.76) \\
    13    & \hspace{1.2 em} Major thoroughfares (50/46298)    & 57.46(1.41) & 55.18(10.1) & \textbf{61.11(2.52)} & 34.23(4.68) & 54.58(12.8) & 23.91(6.15) & 56.74(2.10) \\
    14    & \hspace{5.0 em} Highways (50/9815)                & 53.51(10.6) & 85.65(5.48) & 91.63(1.85) & 89.82(3.57) & \textbf{96.33(2.03)} & 81.11(0.10) & 95.72(1.89) \\
    15    & \hspace{5.4 em} Railways (50/6887)                & 54.39(14.0) & 88.00(3.77) & 92.52(3.63) & 96.54(1.06) & \textbf{99.05(1.14)} & 90.31(2.51) & 98.41(0.63) \\
    16    & \hspace{2 em} Paved parking lots (50/11450)       & 75.32(4.16) & 52.48(8.96) & 77.00(7.93) & 12.56(11.9) & 91.46(3.27) & 80.81(0.61) & \textbf{94.78(0.55)} \\
    17    & \hspace{-0.7 em} Unpaved parking lots (2/144)     & 42.05(24.7) & -     & 52.50(16.2) & -     & 37.81(26.4) & -     & \textbf{75.94(9.61)} \\
    18    & \hspace{7.3 em} Cars (50/6497)                    & 70.23(2.43)          & 76.45(11.1) & 74.92(7.65) & 92.86(4.47) & \textbf{97.92(1.23)} & 33.35(14.0) & 92.76(1.47) \\
    19    & \hspace{6.6 em} Trains (50/5319)                  & 71.17(16.8) & 77.87(5.18) & 95.09(4.20) & 98.01(1.49) & \textbf{99.08(0.90)} & 71.81(3.13) & 96.27(1.92) \\
    20    & \hspace{3.5 em} Stadium seats (50/6774)            & 24.84(7.51) & 99.73(0.27) & \textbf{99.97(0.05)} & -     & 99.71(0.52) & 81.89(4.31) & 99.31(0.40) \\
    \midrule
    \multicolumn{2}{c}{OA} & 71.18(1.82) & 68.67(1.96) & 76.46(1.00) & 57.20(0.09) & 74.69(1.74) & 65.19(1.41) & \textbf{77.36(0.56)} \\
    \multicolumn{2}{c}{AA} & 60.40(2.40) & 55.95(0.74) & 70.22(1.08) & 47.41(0.17) & 68.29(1.95) & 67.16(2.88) & \textbf{71.56(0.63)} \\
    \multicolumn{2}{c}{Kappa} & 64.82(1.83) & 60.91(1.92) & 76.75(0.98) & 46.80(0.64) & 76.00(2.45) & 60.84(1.61) & \textbf{82.39(0.55)} \\
    \hdashline
     \multicolumn{2}{c}{Training Time (s)}    & \multicolumn{1}{r}{137.35}  & \multicolumn{1}{r}{74.69}  & \multicolumn{1}{r}{374.15}  & \multicolumn{1}{r}{984.38}  & \multicolumn{1}{r}{144.89}  & \multicolumn{1}{r}{620.84}  & \multicolumn{1}{r}{200.06} \hspace{0em}  \\
    \multicolumn{2}{c}{Test Time (s)}         & \multicolumn{1}{r}{224.56}  & \multicolumn{1}{r}{41.01}  & \multicolumn{1}{r}{27.73}   & \multicolumn{1}{r}{86.48}   & \multicolumn{1}{r}{107.68}  & \multicolumn{1}{r}{230.13} & \multicolumn{1}{r}{66.94} \hspace{0em} \\ 
    \bottomrule
    \end{tabular}
  \label{tab:Houston2018dataset}}
\end{table*}

\subsubsection{Implementation Details} 
\label{Implementdetail}
The experimental environments for all networks in this paper are configured on a desktop equipped with an Intel i7-10700F processor and a single NVIDIA 2070 Super GPU. The learning rate is 0.001 and $\alpha$ is set to 0.2. 
The MAHiDFNet is developed using the Keras framework. The NNCNet, CALC, AM$^3$Net, MBFormer, DSHFNet, and FusDreamer are constructed using the PyTorch framework. The designed prompts of other datasets are shown in the official repository. 

\subsection{Experimental Analysis}

{To demonstrate the superiority of the proposed FusDreamer, this paper compares the classification performance of our network with six SOTA methods, namely: a multiattentive hierarchical dense fusion
based network (MAHiDFNet) \cite{wang2022multi}, an adaptive mutual
learning-based multimodal network (AM$^3$Net) \cite{wang2022am3net}, a nearest neighbor-based contrastive learning nework (NNCNet) \cite{wang2023nearest}, a coupled adversarial learning for fusion classification (CALC) \cite{lu2023coupled}, a mutually beneficial transformer (MBFormer) \cite{wang2023mutually}, and a dynamic scale hierarchical fusion network (DSHFNet) \cite{10238743}.} This paper leverages these well-established benchmarks to provide a robust comparison, emphasizing the strengths and potential advantages of the proposed approach.

\begin{table*}
  \centering
  {\caption{Quantitative Performance (in \%) of Different Classification Networks on the MUUFL Dataset. The Number in Parentheses Indicates the Standard Variance for the 20 Repeated Experiments, and the Best Classification Results are Shown in Bold.}
    \begin{tabular}{ccccccccc}
     \toprule
    \multicolumn{1}{c}{No.} & Class(Train/Test)  & MAHiDFNet & AM3Net & NNCNet & CALC & MBFormer & DSHFNet & FusDreamer \\
     \midrule
    \multicolumn{1}{c}{1}   & \hspace{8 em} Tree (15/23231)                        & \textbf{93.09(2.53)} & 50.50(7.78) & 81.16(2.56) & 85.06(7.46) & 69.59(5.63) & 85.35(0.51) & 85.39(8.59) \\
    \multicolumn{1}{c}{2}   & \hspace{-2 em} Mostly-grass ground surface (15/4255) & 54.87(6.85) & 40.27(5.31) & 51.74(12.6) & 54.32(21.6) & 37.36(6.62) & \textbf{82.90(7.09)} & {66.00(10.3)} \\
    \multicolumn{1}{c}{3}   & \hspace{1.3 em} Mix ground surface (15/6867)         & \textbf{55.82(8.54)} & 24.85(9.62) & 41.14(3.95) & 52.45(16.8) & 47.32(4.68) & 47.72(14.0) & 39.81(15.4) \\
    \multicolumn{1}{c}{4}   & \hspace{3.8 em} Dirt and sand (15/1811)              & 44.58(10.1) & 65.01(5.24) & 70.43(2.61) & \textbf{87.59(6.42)} & 63.87(10.5) & 86.29(4.99) & 78.04(9.28) \\
    \multicolumn{1}{c}{5}   & \hspace{7.3 em} Road (15/6672)                       & \textbf{95.64(0.80)} & 45.10(5.55) & 67.48(3.92) & 65.42(12.8) & 51.86(9.06) & 76.47(5.58) & 76.10(5.61) \\
    \multicolumn{1}{c}{6}   & \hspace{6.6 em} Water (15/451)                       & 14.21(3.29) & 99.57(0.59) & \textbf{100.0(0.00)} & 99.73(0.60) & 99.72(0.62) & {85.42(4.62)} & 99.95(0.12) \\
    \multicolumn{1}{c}{7}   & \hspace{1.1 em} Shadow of buildings (15/2218)        & 35.40(4.02) & 49.70(3.75) & 81.06(3.42) & 78.59(12.7) & 67.44(8.08) & \textbf{84.00(0.60)} & 80.41(16.1) \\
    \multicolumn{1}{c}{8}   & \hspace{6 em} Building (15/6235)                     & 88.35(10.3) & 76.13(5.53) & 68.28(8.11) & 88.17(7.37) & 71.05(5.03) & 82.29(4.37) & \textbf{89.51(3.11)} \\
    \multicolumn{1}{c}{9}   & \hspace{5.8 em} Sidewalk (15/1370)                   & 36.58(4.71) & 29.66(7.31) & 42.95(6.41) & 37.25(22.7) & 39.63(14.4) & \textbf{59.56(7.08)} & 47.25(10.9) \\
    \multicolumn{1}{c}{10}  & \hspace{4 em} Yellow curb (15/168)                   & 3.86(1.17) & 67.97(12.2) & 58.45(11.9) & 61.13(12.5) & \textbf{76.08(10.6)} & 64.40(4.38) & 51.54(11.4) \\
    \multicolumn{1}{c}{11}  & \hspace{3.8 em} Cloth panels (15/254)                & 17.90(6.49) & 95.40(0.70) & 84.25(7.31) & \textbf{97.66(1.89)} & 97.09(1.49) & 80.43(1.24) & 97.07(1.65) \\
     \midrule
    \multicolumn{2}{c}{OA} & 65.68(2.28) & 49.18(2.83) & 69.24(1.50) & 75.06(3.95) & 61.47(3.80) & 75.81(3.04) & \textbf{75.96(5.04)} \\
    \multicolumn{2}{c}{AA} & 49.12(2.56) & 58.56(0.80) & 60.73(1.69) & 68.02(4.86) & 52.58(3.86) & \textbf{75.43(1.71)} & 71.22(6.22) \\
    \multicolumn{2}{c}{Kappa} & 56.94(2.44) & 38.49(2.62) & 67.90(1.73) & 73.40(3.98) & 65.55(1.12) & 73.21(3.72) & \textbf{73.73(4.66)} \\
    \hdashline
  \multicolumn{2}{c}{Training Time (s)}    & \multicolumn{1}{r}{104.39}  & \multicolumn{1}{r}{44.60}  & \multicolumn{1}{r}{272.63}  & \multicolumn{1}{r}{808.94}  & \multicolumn{1}{r}{27.57}  & \multicolumn{1}{r}{395.39} & \multicolumn{1}{r}{61.40}\hspace{0em}  \\
  \multicolumn{2}{c}{Test Time (s)}       & \multicolumn{1}{r}{27.97}   & \multicolumn{1}{r}{4.85}   & \multicolumn{1}{r}{2.80}  & \multicolumn{1}{r}{74.85}  & \multicolumn{1}{r}{11.11}   & \multicolumn{1}{r}{74.55}   & \multicolumn{1}{r}{18.27}\hspace{0em}  \\
    \bottomrule
    \end{tabular}
  \label{tab:MUUFLdataset}}
\end{table*}

\begin{figure}
  \centering{
  {\includegraphics[scale=0.113]{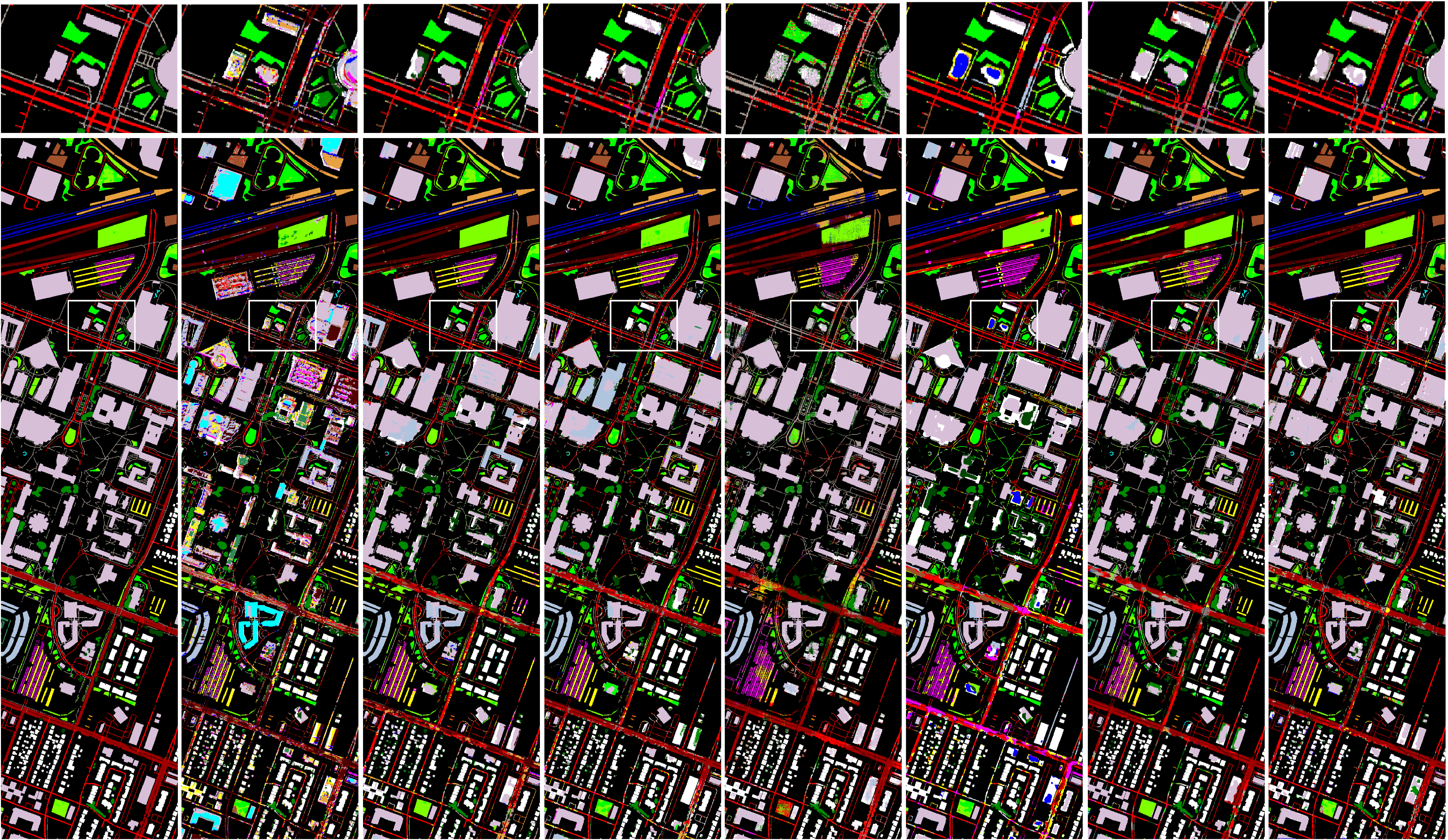}}}\\
  (a) \ \  \ \  \  (b)\ \  \ \  \  (c)\ \  \ \  \ \  (d)\ \ \ \  \ \  (e)\ \ \ \   \ \ (f)\ \ \ \  \   (g)\ \  \ \  \  (h)
  \caption{Classification maps obtained by different networks on the Houston 2018 dataset. (a) Ground Truth, (b) MAHiDFNet, (c) AM$^3$Net, (d) NNCNet, (e) CALC, (f) MBFormer, (g) DSHFNet, and (h) FusDreamer.}
  \label{fig:hu2018small}
\end{figure}

{\subsubsection{Houston 2013 Dataset}Table \ref{tab:Houston2013small} shows the classification performance of different networks on the Houston 2013 dataset. Here, only a small number of samples are used as training samples. As can be observed from Table \ref{tab:Houston2013small}, the proposed method can achieve the highest classification accuracies when compared with other SOTA methods. For instance, the proposed method can achieve the top classification accuracy in OA metric (89.24\%), which is more than 23 points higher than  AM$^3$Net when only using a small number of training samples. Moreover, the proposed method is still 3 points higher than those methods in terms of OA, AA, and Kappa metrics, such as the NNCNet, CALC, MBFormer, and DSHFNet, etc. Meanwhile, the bottom of Table \ref{tab:Houston2013small} presents the runtime of different methods. As seen, although the proposed method does not outperform the AM$^3$Net and MBFormer on training or test time, it performs comparably with the MAHiDFNet in test time and outperforms the CALC in terms of training and test times. As to reasons, the diffusion needs to perform a large number of matrix operations and convolutions during each iteration step, which is time-consuming in high-dimensional space for HSI-LiDAR data. Additionally, mapping prompt-multimodality data into a joint embedding space to extract and align high-dimensional features also demands certain computations.}

\begin{table*}
  \centering
   { \caption{Quantitative Performance (in \%) of Different Classification Networks on the Trento Dataset. The Number in Parentheses Indicates the Standard Variance for the 20 Repeated Experiments, and the Best Classification Results are Shown in Bold.}
  \footnotesize
    \begin{tabular}{ccccccccc}
    \toprule
    \multicolumn{1}{c}{No.} & Class(Train/Test)  & MAHiDFNet &  AM$^3$Net & NNCNet & CALC & MBFormer & DSHFNet & FusDreamer \\
    \midrule
    1     & Apple trees (13/4021)\ \ \ \ \  & \textbf{98.69(1.03)} & 81.49(10.4) & 98.10(1.26) & 76.04(15.4) & 83.99(4.38) & 82.15(9.18) & 95.21(4.52) \\
    2     & Buildings (13/2890)\ \ \        & 91.68(0.79) & 61.41(6.50) & 76.45(7.93) & \textbf{99.74(0.14)} & 98.00(2.81) & 93.46(3.73) & 95.34(7.19) \\
    3     & Ground (11/468) \               & 93.63(4.05) & 72.73(9.64) & 94.79(4.81) & 92.91(3.78) & 90.67(3.97) & 93.12(0.93) & \textbf{95.62(2.62)} \\
    4     & \ Woods (15/9108)               & 99.95(0.06) & 95.44(6.25) & 97.94(2.54) & \textbf{100.0(0.00)} & 97.67(4.20) & 95.23(1.93) & 99.90(0.13) \\
    5     & Vineyard (18/10483)             & 94.24(4.46) & 91.42(6.67) & 96.20(2.54) & 99.96(0.07) & 95.09(0.74) & \textbf{100.0(0.00)} & 99.74(0.36) \\
    6     & \ \ Roads (12/3162)             & \textbf{90.88(3.74)} & 46.51(15.1) & 83.26(6.55) & 86.30(2.83) & 84.43(6.72) & 81.44(3.69) & 77.05(13.0) \\
    \midrule
    \multicolumn{2}{c}{OA} & 95.78(1.84) & 83.57(1.95) & 93.71(0.67) & 95.22(2.38) & 93.51(1.53) & 93.61(1.43) & \textbf{96.36(0.56)} \\
    \multicolumn{2}{c}{AA} & 94.84(1.27) & 74.83(2.76) & 91.60(0.87) & 93.56(3.24) & 91.41(1.99) & 90.90(1.85) & \textbf{95.12(0.76)} \\
    \multicolumn{2}{c}{Kappa} & \textbf{94.32(2.50)} & 78.19(2.22) & 91.12(1.07) & 92.49(3.64) & 91.64(1.29) & 91.12(1.91) & 93.81(1.09) \\
    \hdashline
    \multicolumn{2}{c}{Training Time (s)}  & \multicolumn{1}{r}{60.13}  &\multicolumn{1}{r}{22.45}  & \multicolumn{1}{r}{314.41}  & \multicolumn{1}{r}{225.91}  & \multicolumn{1}{r}{28.77}   & \multicolumn{1}{r}{380.84}  & \multicolumn{1}{r}{53.57}  \\
    \multicolumn{2}{c}{Test Time (s)}       & \multicolumn{1}{r}{17.89}  &\multicolumn{1}{r}{2.39} &   \multicolumn{1}{r}{16.30}  & \multicolumn{1}{r}{48.31}  & \multicolumn{1}{r}{64.60}    & \multicolumn{1}{r}{46.08} & \multicolumn{1}{r}{16.42}  \\
    \bottomrule
    \end{tabular}
  \label{tab:Trentodataset}}
\end{table*}

\subsubsection{Houston 2018 Dataset}Table \ref{tab:Houston2018dataset} shows the classification performance of different networks on the Houston 2018 dataset. Compared with the Houston 2013 dataset, the Houston 2018 dataset contains more classification categories, which means the algorithm needs stronger discriminative capabilities to differentiate between various land cover types. As seen from Table \ref{tab:Houston2018dataset}, the proposed method still has the highest classification performance compared with others, showing the algorithm's robustness and precision in different classification datasets. The classification maps for all networks on the Houston 2018 dataset are established in Fig. \ref{fig:hu2018small}. As seen, the proposed method accurately delineates boundaries between classes, producing sharp and well-defined edges and obtaining better classification performance. 

\subsubsection{MUUFL Dataset}
\begin{figure}[ht]
  \centering{
  {\includegraphics[scale=0.4]{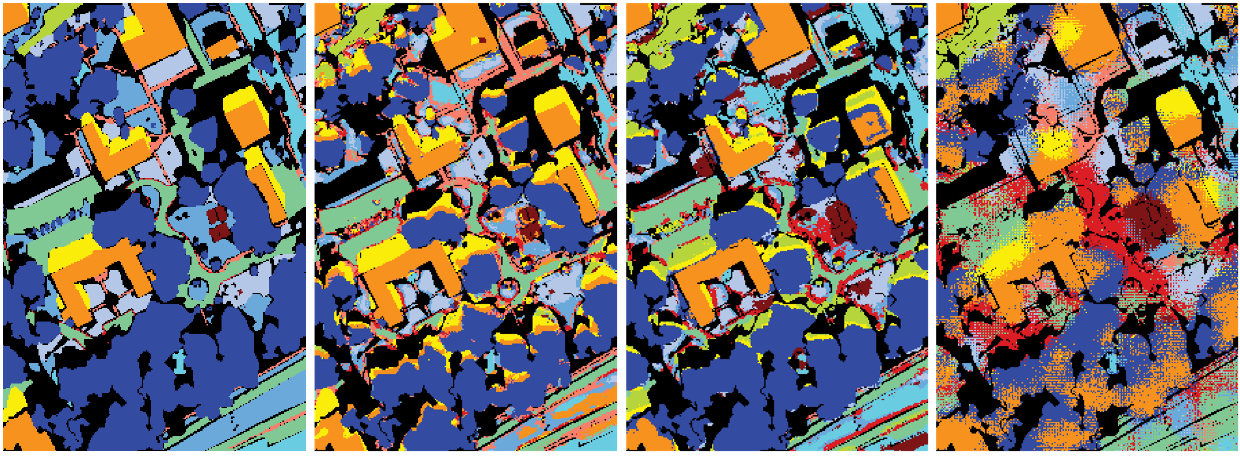}}\\
  (a) \hspace{4em} (b) \hspace{4em} (c) \hspace{4em} (d)\\
  \ {\includegraphics[scale=0.4]{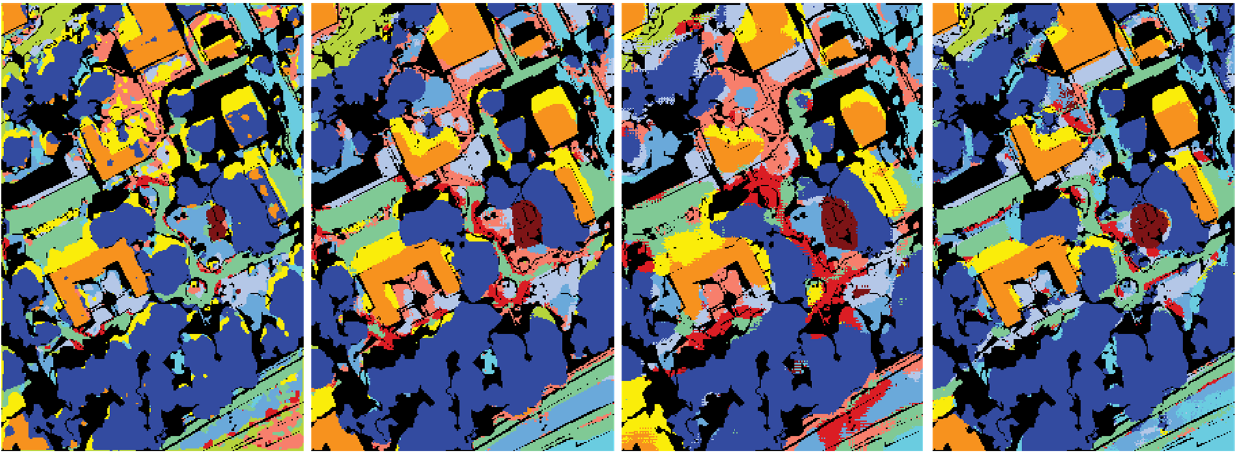}}\\
  \  (e) \hspace{4em} (f) \hspace{4em} (g) \hspace{4em} (h)
  }
  \caption{Classification maps obtained by different networks on the MUUFL dataset. (a) Ground Truth, (b) MAHiDFNet, (c) AM$^3$Net, (d) NNCNet, (e) CALC, (f) MBFormer, (g) DSHFNet, and (h) FusDreamer.}
  \label{fig:muuflsmall}
\end{figure}

Moreover, experiments are conducted on the MUUFL dataset for complex scenario verification. The MUUFL dataset contains more data noise, and the spectral features may be similar, increasing the classification difficulty. Table \ref{tab:MUUFLdataset} shows the classification performance on the  MUUFL dataset. As observed, the MBFormer achieves only 61.47\% classification accuracy when using 15 training samples per class on the MUUFL dataset. MBFormer requires a large number of labeled samples for effective training due to its transformer structure. In contrast, the proposed method leverages a language-multimodal generation model, which has a stronger capability to capture and generate texture structures even with limited samples. Fig. \ref{fig:muuflsmall} shows the classification maps of different methods. As seen, the proposed method achieves the highest classification performance, highlighting its robustness and effectiveness across a diverse range of classes and conditions. 

\subsubsection{Trento Dataset}
\begin{figure}
   \centering{
   {\includegraphics[scale=0.277]{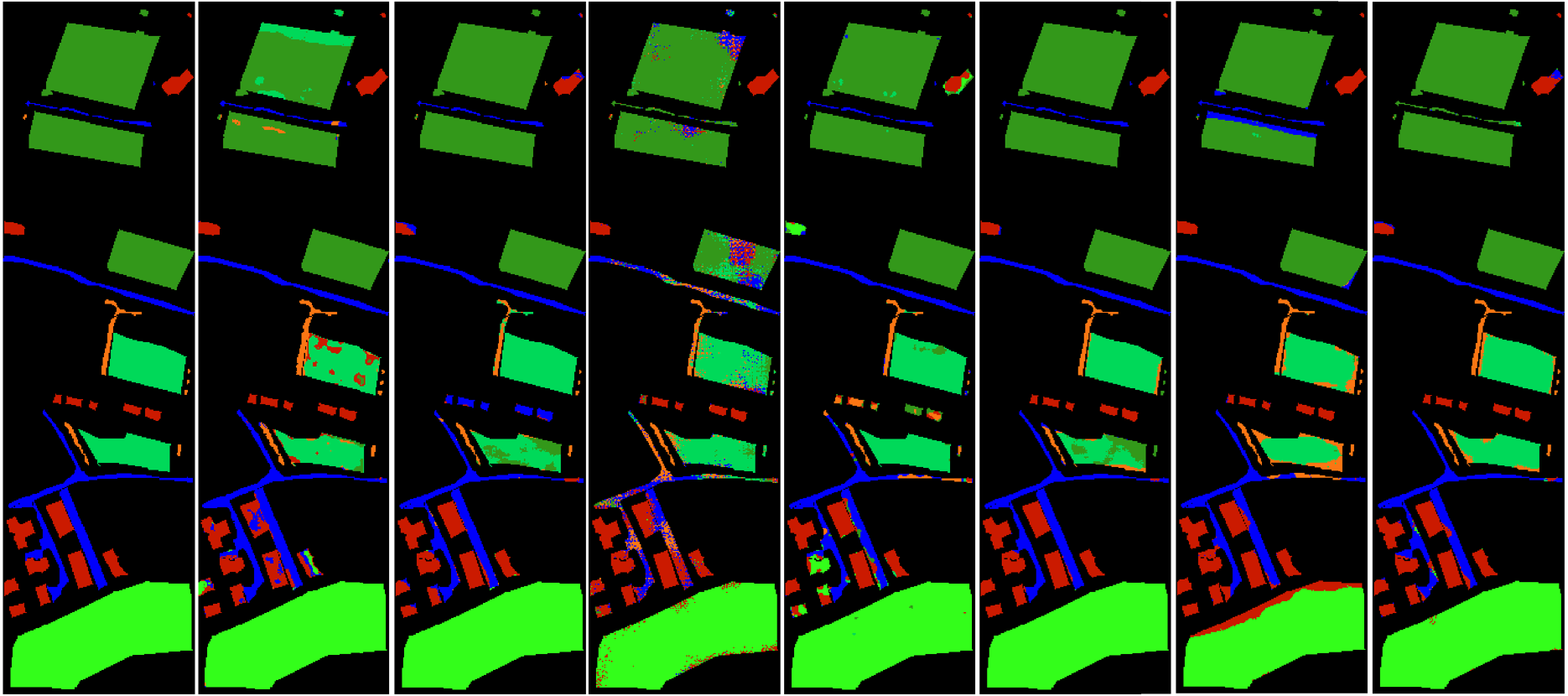}}}\\
  (a) \ \  \ \  \  (b)\ \  \ \  \  (c)\ \  \ \  \ \  (d)\ \ \ \  \ \  (e)\ \ \ \   \ \ (f)\ \ \ \  \   (g)\ \  \ \  \  (h)
 \caption{Classification maps obtained by different networks on the Trento dataset. (a) Ground Truth, (b) MAHiDFNet, (c) AM$^3$Net, (d) NNCNet, (e) CALC, (f) MBFormer, (g) DSHFNet, and (h) FusDreamer.}
   \label{fig:trentosmall}
\end{figure}

Additionally, more experiments are conducted on the Trento dataset for analysis, whose comparison classification results and maps can be seen in Table \ref{tab:Trentodataset} and Fig. \ref{fig:trentosmall}. A similar conclusion can be drawn from Table \ref{tab:Trentodataset}. As seen, the proposed algorithm in this paper significantly outperforms other advanced algorithms in classification (with 97.57\% classification accuracy). It achieves high classification accuracy by effectively capturing differentiated physical details and subtle variations within each class. Fig. \ref{fig:trentosmall} shows a clear and distinct separation of different regions, even in complex areas, demonstrating the algorithm's robustness and precision. 
To sum up, the experimental results indicate that the proposed algorithm has significant advantages in classification tasks. Especially in cases with limited data, the FusDreamer demonstrates high practical value and promising application prospects.

\begin{table}
  \centering
\caption{Performance Comparison with Other SOTA Networks on the Trento Dataset with a Larger Number of Training Samples.}
    \begin{tabular}{ccccc}
    \toprule
    \multicolumn{2}{c}{Methods} & OA    & AA    & Kappa \\
    \midrule
    \multicolumn{1}{r}{A$^3$CLNN\cite{9234528}}      & \multicolumn{1}{l}{IEEE TNNLS' 2022} & 90.55 & 91.81 & 89.75 \\
    \multicolumn{1}{r}{Sal$^2$RN\cite{li2022sal2rn}} & \multicolumn{1}{l}{IEEE TGRS' 2022}  & 98.80 & 97.54 & 98.39 \\
    \multicolumn{1}{r}{DHViT\cite{9755059}}          & \multicolumn{1}{l}{IEEE TIP' 2022}   & 99.55 & 99.61 & 99.51 \\
    \midrule
    \multicolumn{1}{r}{MACRMoI-N\cite{10105921}}     & \multicolumn{1}{l}{IEEE TCSVT' 2023} & 97.47 & 96.29 & 96.65 \\
    \multicolumn{1}{r}{ExViT\cite{yao2023extended}}  & \multicolumn{1}{l}{IEEE TGRS' 2023}  & 98.58 & 94.53 & 98.10 \\
    \multicolumn{1}{r}{SOT-Net\cite{9775021}}        & \multicolumn{1}{l}{IEEE TCYB' 2023}  & 99.14 & 98.49 & 98.86 \\
    \midrule
    \multicolumn{1}{r}{S2EFT\cite{FENG2024111190}}   & \multicolumn{1}{l}{KBS' 2024}        & 98.45 & 97.62 & 97.82 \\
    \multicolumn{1}{r}{S2LFNet\cite{cao2023spectral}}       & \multicolumn{1}{l}{IEEE TGRS' 2024}  & 98.82 & 97.31 & 98.23 \\
    \multicolumn{1}{r}{GAMF\cite{cai2024novel}}      & \multicolumn{1}{l}{ESWA' 2024}       & 98.96 & 97.47 & 98.61 \\
    \multicolumn{1}{r}{CMSE\cite{10443842}}          & \multicolumn{1}{l}{IEEE TGRS' 2024}  & 99.08 & 98.71 & 98.77 \\
    \multicolumn{1}{r}{SMDN\cite{10609507}}          & \multicolumn{1}{l}{IEEE TBD' 2024}   & 99.18 & 98.68 & 98.97 \\
    \multicolumn{1}{r}{UCAFNet\cite{10595999}}       & \multicolumn{1}{l}{IEEE TGRS' 2024}  & 99.26 & 98.76 & 99.01 \\
    \multicolumn{1}{r}{MIViT\cite{10464367}}         & \multicolumn{1}{l}{IEEE TCSVT' 2024} & 99.43 & 99.16 & 99.24 \\
    \midrule
    \multicolumn{2}{c}{FusDreamer ($e=3$)}                                                  & 99.27 & 99.21 & 99.44 \\
    \multicolumn{2}{c}{FusDreamer ($e=5$)}                                                  & \textbf{99.59} & \textbf{99.68} & \textbf{99.57} \\
    \bottomrule
    \end{tabular}
  \label{tab:SOTAcomaprision}
\end{table}

\subsubsection{Comparison with Other SOTA methods} Table \ref{tab:SOTAcomaprision} presents a more detailed comparison of SOTA models developed over the past two years on the Trento dataset, with all results directly cited from the original papers and the training samples following Ref. \cite{wang2022am3net}. Here, the following thirteen networks, including the A$^3$CLNN \cite{9234528}, al$^2$RN \cite{li2022sal2rn}, DHViT \cite{9755059}, MACRMoI-N \cite{10105921}, ExViT \cite{yao2023extended}, SOT-Net \cite{9775021}, S2EFT \cite{FENG2024111190}, S2LFNet \cite{cao2023spectral}, 
GAMF \cite{cai2024novel}, CMSE \cite{10443842}, SMDN \cite{10609507}, UCAFNet \cite{10595999}, and MIViT \cite{10464367}, are used for comparison. As seen from Table \ref{tab:SOTAcomaprision}, the proposed FusDreamer can always yield the highest classification performance based on three or five transformer blocks. This indicates its effectiveness in handling diverse and complex data scenarios, making it a reliable choice for various remote sensing classification tasks. The detailed experimental results with a larger number of training samples on four datasets are shown in the official repository.

\begin{table}
  \centering
  \caption{Classification Performance with Different Fusion Strategies on the Trento Dataset.}
  \footnotesize
  \setlength{\tabcolsep}{5.6mm}
    {\begin{tabular}{lccc}
    \toprule
    Diffusion Weights & OA    & AA    & Kappa \\
    \midrule
    No Fusion           & 93.34 & 91.15 & 89.28 \\
    Sum                 & 95.72 & 94.29 & 93.38 \\
    Concat              & 96.42 & 95.21 & 94.21 \\
    Weighted Sum        & 96.93 & 95.89 & 93.23 \\
    Weighted Concat     & 97.11 & 96.01 & 94.03 \\
    The proposed RDAF   & \textbf{97.57} & \textbf{96.74} & \textbf{95.13} \\
    \bottomrule
    \end{tabular}}
  \label{tab:fusionstrategies}
\end{table}

\subsection{Ablation Studies} 
To validate the effectiveness of different components in the model design process and assess the contribution of each module within the algorithm, a series of ablation experiments are conducted as below.


\subsubsection{The Impact of Dimension and Patch Size}
\begin{figure}
   \centering{
   {\includegraphics[scale=1]{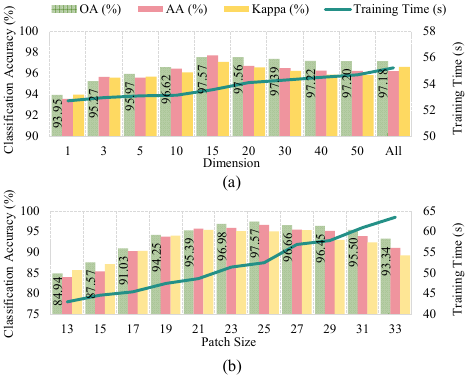}}}
 \caption{The impacts of spectral dimension and patch size based on the proposed method for the Houston dataset.}
   \label{fig:dimensionandpatchsize}
\end{figure}

Fig. \ref{fig:dimensionandpatchsize} shows the classification performance of the FusDreamer under different spectral dimensions and different patch sizes on the Houston dataset. As seen from Fig. \ref{fig:dimensionandpatchsize} (a), the OA, AA, and Kappa metrics initially improve with increasing spectral dimensions and then stabilize. In contrast, the training time shows near-linear growth with increasing channel dimensions. To balance classification performance and training time, this paper selects 15 as the default parameter. As seen from Fig. \ref{fig:dimensionandpatchsize} (b), different patch sizes directly influence feature representation and classification accuracy, with a notable change observed at a patch size of 25 and a sharp increase in training time beyond this point. Therefore, the patch size is set to 25 as the default
parameter on four datasets.

\subsubsection{The Impact of Fusion Strategies}
Table \ref{tab:fusionstrategies} indicates the classification performance of the proposed method under fusion strategies in the diffusion module on the Trento dataset. Specifically, there are six different fusion strategies during the diffusion module, \emph{i.e.}, no fusion, sum, concatenation, weighted sum, weighted concatenation, and the proposed RDAF. As can be seen from Table \ref{tab:fusionstrategies}, compared to direct addition or stacking fusion methods, an adaptive fusion strategy usually offers greater flexibility and precision in the fusion process. When using weighted sum and concat fusion strategies, they can respectively obtain 96.93\% and 97.11\% classification performance. In contrast, this paper adopts the RDAF method for fusion, achieving the highest classification accuracy at 97.57\%. This indicates that RDAF effectively leverages the advantages of each branch feature while avoiding information or feature redundancy, thereby enhancing the overall model's performance and accuracy.

\begin{table}
  \centering
 {  \caption{The classification performance of the FusDreamer under different prompt design on the Trento dataset.}
    \begin{tabular}{cccc}
    \toprule
    \multicolumn{1}{c}{Desiged Pormpts} & OA    & AA    & Kappa \\
    \midrule
    \multicolumn{1}{c}{with inaccurate self-categorical descriptions} & 89.81 & 87.03 & 88.42 \\
    \multicolumn{1}{c}{without self-categorical descriptions} & 94.79 & 94.01 & 94.13 \\
    \multicolumn{1}{c}{without adjacency relationships} & 94.93 & 94.11 & 94.38 \\
    with well-designed prompts & \textbf{97.57} & \textbf{96.74} & \textbf{95.13} \\
    \bottomrule
    \end{tabular}%
  \label{tab:prompts}}
\end{table}%

{\subsubsection{The Impact of the Prompts}
Table \ref{tab:prompts} illustrates the classification performance of FusDreamer under different prompt designs on the Trento dataset. As shown in the first row, using inaccurate label descriptions results in a significant drop in performance, with accuracy decreasing from 97.57\% to 89.81\%. This indicates that poorly designed prompts, which lack semantic alignment with the dataset labels, can lead to misclassification. Misaligned prompts may misrepresent the dataset’s semantic structure, thereby severely undermining the classifier's ability to make accurate predictions. The absence of self-categorical and adjacency relationship descriptions causes a slight drop in performance, emphasizing their importance in helping the model discern subtle distinctions between categories. Self-categorical descriptions likely provide the model with contextual information about each class, while adjacency relationships help the model account for dependencies and interactions between classes. In contrast, the proposed prompts consistently deliver the best results across all metrics, demonstrating the effectiveness of the prompt design strategy. }

\begin{table}
  \centering
  \caption{Classification Performance Inclusion or Exclusion Each Component on the Trento Dataset.}
  \footnotesize
  \setlength{\tabcolsep}{3.5mm}
    {\begin{tabular}{cccccc}
    \toprule
    {\multirow{2}[0]{*}{LaMG}}        &  \multicolumn{2}{c}{OK-CP}&{\multirow{2}[0]{*}{OA}} &{\multirow{2}[0]{*}{AA}} &{\multirow{2}[0]{*}{Kappa}} \\
       & MFE  & MPE  &     &     &  \\
    \midrule
      \checkmark    &                 &                 &  86.72 & 82.68 & 80.34\\
                    &  \checkmark     &  \checkmark     &  90.65 & 90.20 & 90.45 \\
     \checkmark     &  \checkmark     &                 &  95.78 & 94.71 & 94.82 \\
     \checkmark     &                 &  \checkmark     &  93.76 & 93.03 & 93.35 \\
     \checkmark     &  \checkmark     &  \checkmark     & \textbf{97.57} & \textbf{96.74} & \textbf{95.13} \\
    \bottomrule
    \end{tabular}}
  \label{tab:differentmodules}%
\end{table}%

\subsubsection{The Impact of Each Component} 
Table \ref{tab:differentmodules} verifies the classification performance for studying the effectiveness of different components, \emph{i.e.}, the LaMG (Sec. \ref{DAFM}), OK-CP (Sec. \ref{TMP}) and its subsection, on the Trento dataset. Each module is individually removed, and the classification accuracies are evaluated in Table \ref{tab:differentmodules}. As observed from the first and third lines in Table \ref{tab:differentmodules}, the removal of the LaMG or OK-CP modules module led to a massive drop in accuracy, indicating its crucial role in latent diffusion or open-world knowledge-guided components in capturing generative macro-world relevant features. The proposed method achieves only 86.72\% performance without the OK-CP module, indicating that pre-trained open-world knowledge facilitates domain-invariant learning, especially with limited labels. For more details, the absence of the feature encoders, \textit{i.e.}, the MFE and MPE, results in significant performance degradation, suggesting the importance of the proposed method's multimodel and multi-attribute prompts encoder structures in generating the world model's features.

\begin{table}
  \centering
  \caption{Classification Performance with Different Loss Functions on the Trento Dataset.}
  \footnotesize
  \setlength{\tabcolsep}{3.9mm}
    {\begin{tabular}{cccccc}
    \toprule
    $\mathscr{Loss}_C$ & $\mathscr{Loss}_N$ & $\mathscr{Loss}_{M}$ & OA    & AA    & Kappa \\
    \midrule
    \checkmark     &                &                & 93.61 & 93.54 & 92.99 \\
    \checkmark     & \checkmark     &                & 95.88 & 94.94 & 93.83 \\
    \checkmark     &                & \checkmark     & 96.36 & 95.12 & 94.81 \\
    \checkmark     & \checkmark     & \checkmark     & \textbf{97.57} & \textbf{96.74} & \textbf{95.13} \\
    \bottomrule
    \end{tabular}}
  \label{tab:loss}
\end{table}


\begin{table}
  \centering
  \caption{Classification Performance with Different Weights of Loss Function on the Trento Dataset.}
  \footnotesize
  \setlength{\tabcolsep}{4.5mm}
    {\begin{tabular}{cccccc}
    \toprule
    $\lambda_1$  & $\lambda_2$  & $\lambda_3$  & OA    & AA    & Kappa \\
    \midrule
    0.2   & 0.4   & 0.4   & 96.05 & 94.73 & 92.60 \\
    0.4   & 0.3   & 0.3   & 97.12 & 96.15 & 94.97 \\
    0.6   & 0.2   & 0.2   & \textbf{97.57} & \textbf{96.74} & \textbf{95.13} \\
    0.8   & 0.1   & 0.1   & 96.61 & 95.48 & 95.48 \\
    \bottomrule
    \end{tabular}}
  \label{tab:weights}
\end{table}

\subsubsection{The Impact of Different Loss Function} 
\label{loss}
Table \ref{tab:loss} shows the classification performance of the proposed method using three different loss functions on the Trento dataset, \emph{i.e.}, the multimodal classification loss $\mathscr{Loss}_{C}$, noise prediction loss $\mathscr{Loss}_{N}$, and text-to-multimodal feature alignment loss $\mathscr{Loss}_{M}$. As seen, when only using $\mathscr{Loss}_{C}$, the model achieves an essential classification accuracy of 93.61\%. By adding more constraints $\mathscr{Loss}_{M}$ and $\mathscr{Loss}_{N}$, not only can the diffusion reverse direction and feature transmission process be directly affected, but also the alignment of prompts with multimodal features is improved. Moreover, when combining three loss functions, the proposed method can obtain the highest and most robust classification result of 97.57\%, demonstrating the importance and advantages of the proposed loss function in distinguishing complex ground objects.

In addition, more experiments are conducted on the Trento dataset to determine the optimal weights for three loss functions: $\mathscr{Loss}_{C}$, $\mathscr{Loss}_{N}$, and $\mathscr{Loss}_{M}$, shown in Table \ref{tab:weights}. To be more specific, the weights $\lambda_1$, $\lambda_2$, and $\lambda_3$ are respectively assigned to each loss function to fine-tune the model's performance. As seen, the proposed FusDreamer can obtain the highest classification accuracy when using $\lambda_1 = 0.6$, $\lambda_2 = 0.2$, and $\lambda_3 = 0.2$, indicating the importance of both $\mathscr{Loss}_{M}$ and $\mathscr{Loss}_{N}$ in constraining feature generation. Therefore, these weights are used as the default configuration for the classification task.

\begin{table}
  \centering
\caption{Classification Performance with Different Number of Transformers on the MPE Module.}
  \footnotesize
  \setlength{\tabcolsep}{2.9mm}
    {\begin{tabular}{cccccc}
    \toprule
     * $e$ & OA    & AA    & Kappa & Training Time & Parameters\\
    \midrule
    * 1   & 95.21 & 93.61 & 95.00 & 38.49 &34,571,227 \\
    * 2   & 96.33 & 95.09 & 94.36 & 45.28 &37,723,611 \\
    * 3   & \textbf{97.57} & \textbf{96.74} & \textbf{95.13} & \textbf{53.57}& 40,875,995\\
    * 4   & 97.89 & 97.16 & 96.40 & 60.31 &44,028,379 \\
    * 5   & 98.34 & 97.78 & 97.16 & 67.44 &47,180,763 \\
    \bottomrule
    \end{tabular}}
  \label{tab:Transformers}
\end{table}

\subsubsection{The Impact of Different Number of Transformers} 
\label{transformernumber}

Table \ref{tab:Transformers} indicates the different classification performances of the proposed method under different numbers of transformers. Generally, the training times and parameters are directly proportional to the number of transformers. Due to the multi-head attention mechanism in the transformer, where each attention head independently computes self-attention, the model's expressive power and classification accuracies are significantly enhanced; however, this also results in increased computational overhead and memory consumption. Therefore, to achieve a balance among classification performance, training time, and model parameters, the number of transformer blocks is set to 3 as the default configuration.

\section{Conclusions}
\label{conclusions}
In this paper, a label-efficient remote sensing world model is developed for hyperspectral and LiDAR data fusion. In contrast to vision-centric models that rely on neural networks to extract specific visual patterns from multimodal data, this paper introduces a world model designed as a unified representation framework to abstract common and high-level knowledge, enabling a latent interaction space across different modality data, i.e., HSI, LiDAR, and text data. Notably, a novel diffusion model is employed to encode visual data, facilitating the generation of high-level features. Additionally, the paper incorporates detailed attribute descriptions grounded in open-world knowledge. Leveraging such a pre-trained VLM model along with domain-specific RS expertise, the proposed FusDreamer achieves promising classification performance, even with limited training data. Experimental results on four real HSI-LiDAR datasets demonstrate that the proposed method outperforms several compared methods in both the visual quality of the classification map and quantitative metrics.\par
{For future work, we plan to investigate the integration of adaptive soft prompts with dynamic contextual learning, aiming to further improve classification performance and enhance generalizability across diverse datasets.}

\bibliographystyle{IEEEtran}
\bibliography{refs}
\end{document}